\documentclass[twocolumn,10pt]{asme2ej}

\usepackage{graphicx} 
\graphicspath{{figure/pdf/}{figure/png/}{figure/jpg/}{figure/eps/}}
\DeclareGraphicsExtensions{.pdf,.png,.jpg,.eps}
\usepackage[caption=false,font=footnotesize]{subfig}
\usepackage{amsmath} 
\usepackage{amssymb}  
\usepackage{mathtools}
\usepackage{todonotes}
\usepackage[binary-units=true]{siunitx}

\title{A Low-Cost, Highly Customizable Solution for Position Estimation
   in Modular Robots}

\author{Chao Liu
    \affiliation{
      Ph.D Student\\
      GRASP Laboratory\\
      Dept. of Mechanical Engineering\\ and Applied Mechanics\\
      University of Pennsylvania\\
      Philadelphia, PA 19104\\
      Email: chaoliu@seas.upenn.edu
    }
}

\author{Tarik Tosun
    \affiliation{
    Ph.D\\
    GRASP Laboratory\\
    Dept. of Mechanical Engineering\\ and Applied Mechanics\\
    University of Pennsylvania\\
    Philadelphia, PA 19104\\
    Email: tarik.d.tosun@gmail.com
    }
}

\author{Mark Yim
  \affiliation{ Professor\\
    GRASP Laboratory\\
    Dept. of Mechanical Engineering and Applied Mechanics\\
    University of Pennsylvania\\
    Philadelphia, PA 19104\\
    Email: yim@seas.upenn.edu
    }
}

\begin{document}

\maketitle

\begin{abstract}
  {\it Accurate position sensing is important for state estimation and
    control in robotics. Reliable and accurate position sensors are
    usually expensive and difficult to customize.  Incorporating them
    into systems that have very tight volume constraints such as
    modular robots are particularly difficult. PaintPots are low-cost,
    reliable, and highly customizable position sensors, but their
    performance is highly dependent on the manufacturing and
    calibration process. This paper presents a Kalman filter with a
    simplified observation model developed to deal with the
    non-linearity issues that result in the use of low-cost
    microcontrollers. In addition, a complete solution for the use of
    PaintPots in a variety of sensing modalities including
    manufacturing, characterization, and estimation is presented for
    an example modular robot, SMORES-EP. This solution can be easily
    adapted to a wide range of applications.}
\end{abstract}

\begin{nomenclature}
\entry{$x$ or $\theta$}{Angular position}
\entry{$u$ or $\omega$}{Angular velocity of the driving motor(s)}
\entry{$V_i$}{Voltage measurement from wiper $i$}
\entry{$\bar{\theta}_i$}{Shifted position measurement from wiper $i$}
\entry{$k$}{Transmission ratio}
\entry{$n$}{Gaussian white noise in transition model}
\entry{$z$}{Measurement in Kalman filter}
\entry{$x^i$}{Reported state from the $i$th feature}
\entry{$z^i$}{Measurement from the $i$th feature}
\entry{$v^i$}{Observation noise of the $i$th feature}
\entry{$N(\bar{\mu}, \bar{\Sigma})$}{Predicted Gaussian distribution of the angular
  position}
\entry{$N(\mu, \Sigma)$}{Updated Gaussian distribution of the angular position}
\entry{$K_t$}{Kalman gain at time $t$}
\end{nomenclature}


\section{Introduction}

It is necessary to have accurate position sensing for most robotics
control processes.  Commercial position sensors are available for many
applications. However, the form factor often presents a challenge to
use these commercial off-the-shelf sensors. This is especially true
for compact robotic systems with tight space constraints such as
modular robotic systems~\cite{modular_robots_system}. Customizable
position sensors give more flexibility to fit these
situations. 

Servos have been used in many modular robotic systems, such as
Molecule~\cite{conro} and CKBot~\cite{ckbot}. Servo motors have
built-in position control circuits. Some modular robotic systems, such
as 3D Fracta~\cite{fracta}, M-TRAN III~\cite{m-tran-3}, and
SuperBot~\cite{superbot}, use rotary potentiometers for position
feedback. Optical or hall-effector sensor encoders are used in
PolyBot~\cite{polybot}, Crystalline~\cite{crystal}, and
ATRON~\cite{atron} to report position information. These devices are
usually too large for highly space-constrained systems, such as
modular robots like SMORES-EP~\cite{ep-face}.

PaintPots are highly-customizable and low-cost position sensors that
can be easily manufactured by widely accessible materials (spray paint
and plastic sheets) and tools (laser cutters or
scissors)~\cite{paintpot}. The sensors can exist in different forms in
terms of size, shape, and surface curvature. Thus, these sensors can be
easily integrated with well designed parts or systems. The sensing
performance and cost of PaintPot sensors make them competitive with
commercial potentiometers~\cite{paintpot} yet the customizability
enable the use in situations which are not possible with commercial
potentiometers.

\begin{figure}[t]
  \centering
  \includegraphics[width=0.4\textwidth]{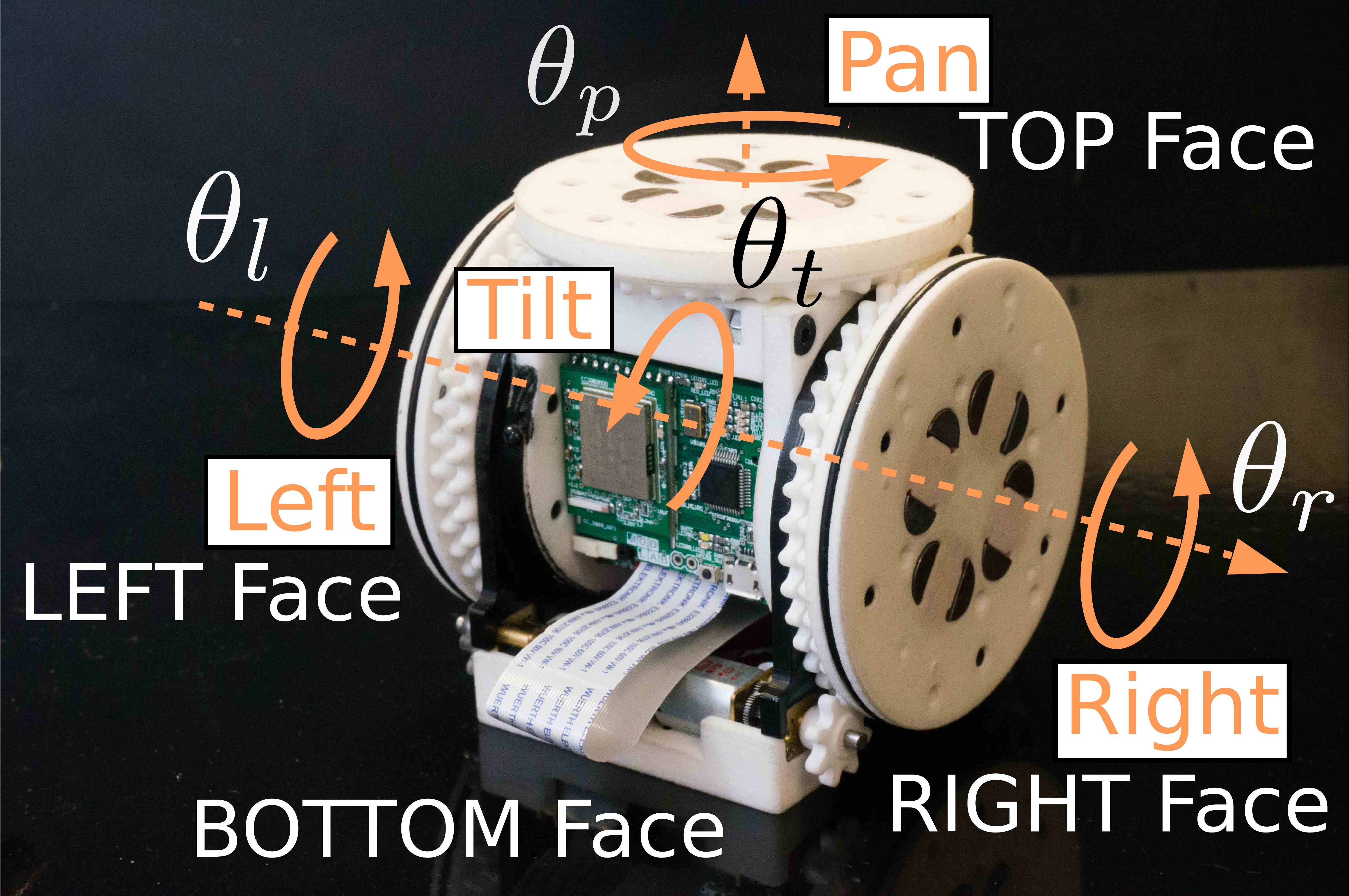}
  \caption{A SMORES-EP module has four degrees of freedom and four connectors.}
  \label{fig:smores-ep}
\end{figure}

Two different designs of PaintPot sensors are used in SMORES-EP
modular robots. In each SMORES-EP module, there are four
degrees of freedom (DOFs) requiring position sensing shown in
Fig.~\ref{fig:smores-ep} --- three continuously rotating joints
(\textit{LEFT DOF}, \textit{RIGHT DOF}, \textit{PAN DOF}) and one
bending joint with a \SI{180}{\degree} range of motion (\textit{TILT
  DOF})~\cite{ep-face}.  Conductive spray paint is used to generate a
resistive track surface. The manufacturing process is easy enough for
a person to make a sensor quickly, but often does not yield consistent
measurement and performance. The terminal-to-terminal resistance can
vary over a large range depending on the thickness of the paint with a
non-linear output. This fact complicates the position estimation for
every DOF.

For state estimation, stochastic techniques based on the probabilistic
assumptions of the uncertainties in the system are widely applied. For
linear systems, the Kalman filter~\cite{kalman} has been shown to be a
reliable approach where uncertain parts in systems are assumed to have
a particular probability distribution, usually Gaussian. Extensions
including the extended Kalman filter (EKF) and unscented Kalman filter
(UKF)~\cite{ukf} have been developed for nonlinear systems. For
SMORES-EP DOF state estimation, we developed a new Kalman filter with
a simpler observation model considering the non-linearity of PaintPot
sensors. A complete and convenient calibration process is developed to
precisely characterize each position sensor quickly. Four estimators
can run on a \SI{72}{\mega\hertz} microcontroller at the same time to
track the states of all DOFs in a SMORES-EP module.

There are many applications that could use potentiometers but are too
size constrained. For example goniometers in instrumented gloves for
virtual reality currently use expensive strain gauge or non-linear
flex sensing technologies. A joint angle potentiometer would be a good
low-cost solution except for size constraints.  Any compact device
with a hinge (a flip phone, smart eyeglasses etc. or wearable devices)
could have that joint angle measured to provide feedback.  This paper
presents an example solution of a class of potentiometers that is
easily adapted to fit many low profile applications. It is semi-custom
in that it is easily scalable to short-run numbers. These sensors are
easily characterized and combined with our state estimation method,
PaintPot sensors can provide reliable position information with little
computation cost. Our solutions in SMORES-EP system shows that
PaintPot sensors are promising for a variety of robotic applications.

The paper is organized as follows. Section~\ref{sec:related} reviews
relevant and previous work. Section~\ref{sec:sensor} introduces the
manufacturing and customized design for SMORES-EP system, as well as
some necessary information. The characterization process is shown in
Section~\ref{sec:characterization} and the estimation method is presented
in Section~\ref{sec:estimate} for all DOFs. Some experiments are shown in
Section~\ref{sec:experiment}. Finally, Section~\ref{sec:conclusion} talks
about the conclusion.

\section{Related Work}
\label{sec:related}

\begin{figure}[t]
  \centering
  \includegraphics[width=0.4\textwidth]{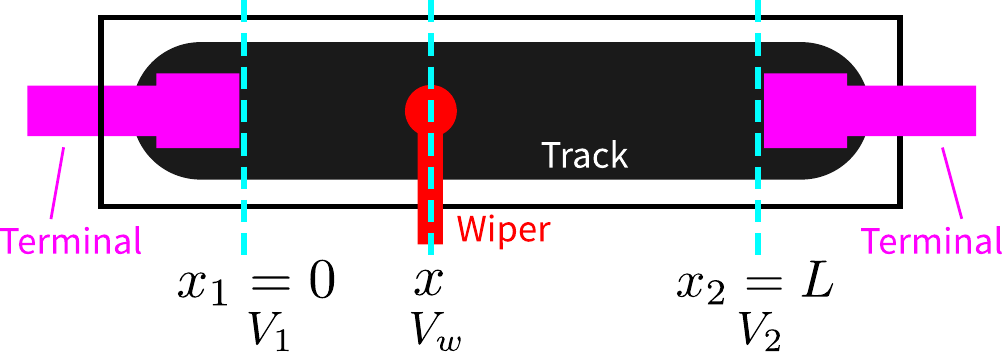}
  \caption{A Potentiometer has three terminals: two fixed electrical
    terminals at the resistive track ends and one electrical contact
    (wiper) that can move along the track surface.}
  \label{fig:potentiometer}
\end{figure}

A potentiometer is a three-terminal resistor with a
sliding or rotating contact (or wiper) that functions as a voltage
divider~\cite{ieee-standards}. There are three basic components in a
potentiometer (Fig.~\ref{fig:potentiometer}): a resistive track, fixed
electrical terminals on the track ends, and an electrical
wiper. Different from most modern potentiometer tracks that are continuous
semiconductive surfaces made of graphite, ceramic-metal composites
(cermets), conductive plastics, or conductive polymer pastes, PaintPot
sensors use an inexpensive carbon-embedded polymer spray
paint~\cite{paintpot}.

Commercial vendors can customize potentiometers using industrial
processes. For example, some manufacturers offer inkjet-printed
thick-films that are deposited on printed circuit boards and other
user specified surfaces and shapes for tracks. Two significant
advantages of PaintPot solutions over these industrial processes are
cost and time: compared with processes that often require thousands of
dollars in up-front engineering fees, produce sensors that cost tens
of dollars each, and have turnaround times of a week or more, PaintPot
solutions can be used by a person or a team to create customized
position sensors with the cost on the order of \SI{1}[\$]{USD} per
unit and also allows rapid iteration (limited only by the drying time
of the paint).

Similar to PaintPot sensors, many rapid prototyping technologies allow
electronics to be integrated into everyday objects at low
cost. Self-folding printable resistors, capacitors, and inductors made
of aluminum-coated polyester film (mylar) sheets are introduced
in~\cite{printable-device}.~\cite{instant-circuit} presents a
technique to accurately print low-resistance traces on sheet materials
using silver nanoparticle ink deposited with a standard inkjet
printer. PaintPot sensors can generate high resistance on order of
\SI{1}{\kilo\ohm} which is desirable for potentiometers used as
voltage dividers.

This paper extends the conference version~\cite{paintpot} on the same
topic that includes the manufacturing process and performance of
PaintPot sensors. PaintPot tracks are hand-painted using spray cans on
\SI{0.79}{mm} thick plastic sheets. The sensors have good performance
in terms of repeatability, resolution, and hysteresis compared with
commercial potentiometers, but have a shorter lifetime. However, the
performance of PaintPot sensors are highly dependent on the
calibration process. This paper extends the conference version by
presenting an integral and fast calibration process to characterize
PaintPot sensors in SMORES-EP. Based on this, a new Kalman filter with
a simplified observation model is developed to increase the
reliability of these position sensors, especially for PaintPot sensors
for continuously rotating joints where there are two coupled electric
contacts, and the performance can be greatly improved compared with
the simple rule presented in~\cite{paintpot}. This reliable
performance allows us to control a SMORES-EP chain to perform
manipulation tasks~\cite{Liu-control-planning-irc2020}.

\section{PaintPot Sensor in SMORES-EP}
\label{sec:sensor}
PaintPots were originally developed to serve as position encoders for
the SMORES-EP modular reconfigurable robot system \cite{paintpot}.
Each SMORES-EP module (Fig.~\ref{fig:smores-ep}) is the size of an
\SI{80}{mm} cube, and has an onboard battery, a microcontroller, and a
Wi-Fi module to send and receive messages, as well as four actuated
joints and electro-permanent magnet connectors on each face, allowing
modules to connect to one another and self-reconfigure
\cite{daudelin2018integrated, smores-reconfig}.  The tight space
requirements of SMORES-EP made it very difficult to incorporate
off-the-shelf position encoders into the design, motivating the
development of custom PaintPot encoders that occupy very little space
within the robot. This section provides an overview of the
manufacturing techniques used to create PaintPots, the fundamental
material resolution of the sensors, and the design of the PaintPots
used in SMORES-EP.  For more detail, we refer the readers to
\cite{paintpot}.

\subsection{Manufacturing Overview}
\begin{figure}[t]
  \centering
  \subfloat[]{\includegraphics[height=0.115\textwidth]{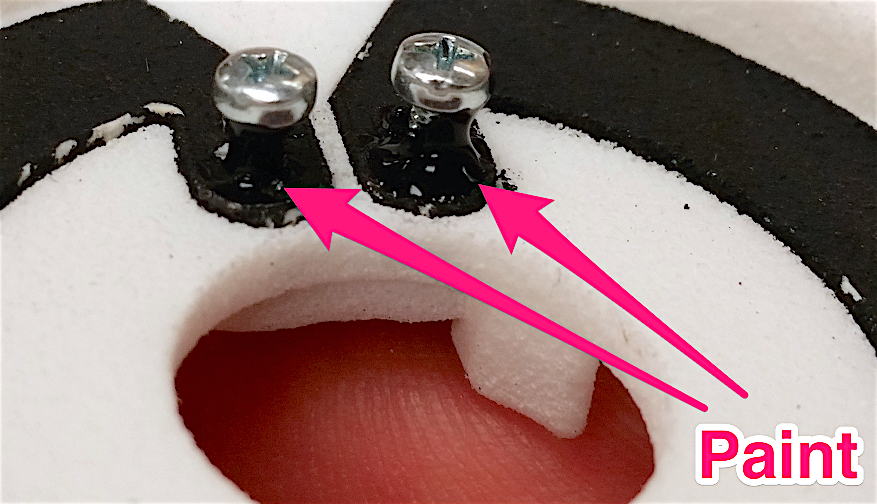}}
  \hfil
  \subfloat[]{\includegraphics[height=0.115\textwidth]{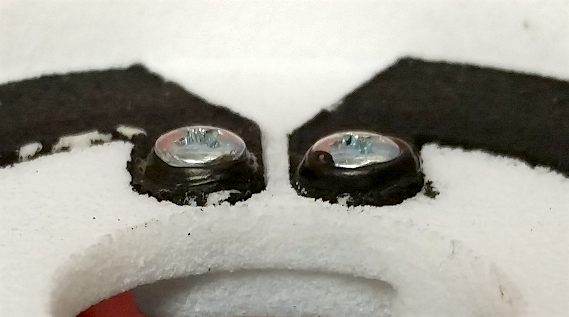}}
  \caption{A bead of conductive paint applied beneath the screw head
    forms a good electrical connection with the track}
  \label{fig:electrical-terminal}
\end{figure}

The resistive track surface of a PaintPot encoder is made from
conductive spray paint on a plastic track substrate. The PaintPots in
SMORES-EP use three coats of MG Chemicals Total Ground conductive
paint \cite{mg-chemicals} sprayed onto Acrylonitrile butadiene styrene
(ABS) plastic sheet. Three coats of paint are applied, with five
minutes of drying time between coats, following the painting
guidelines in the datasheet.  ABS plastic can be cut to precise shapes
in a laser cutter, and forms an ideal substrate for the paint, which
readily bonds to the surface \cite{mg-chemicals}.  Electric terminals
are created by mounting zinc-coated screws at the ends of the
resistive strip. Applying paint above and beneath the screws creates
an electrical connection between the track surface and the screw
(Fig. \ref{fig:electrical-terminal}). Leaded solder adheres to
zinc-coated screws, allowing wires to be attached and detached.

Each PaintPot uses one or more wipers to measure the voltage at the
point of contact with the resistive strip. Wipers with high contact
pressure should be avoided, as they may scratch the paint. Larger
contact surface area also reduces contact resistance, improving signal
quality. The Harwin S1791-42 EMI Shield Finger
Contact~\cite{wiper-datasheet} is used in SMORES-EP. The wiper is a
\SI{4}{mm} high gold-plated tin spring contact with a
\SI{1.45}{mm}$\times$\SI{2.05}{mm} contact area, and a contact force of
\SI{1}{N} (mounted on a PCB at a \SI{3}{mm} working height).%
\subsection{Performance Characteristics}
The fundamental material resolution of the resistive track was
determined by characterizing the signal-to-noise ratio of the measured
voltage on small length scales.  Based on our experiments, the
fundamental resolution of the track material is
$8.63\pm$\SI{0.126}{\mu m}, which is of the same order of magnitude as
some commercially available high-precision potentiometers.  In
SMORES-EP, the limiting factor in resolution is the analog-to-digital
conversion bit depth (\SI{10}{\bit}, or \SI{84}{\mu m}); to reach the
material limit, \SI{14}{\bit} of analog-to-digital conversion depth
would be required.

While PaintPots are not intended for long-lifetime applications, our
analysis found that circular painted tracks have a lifetime of about
50,000 cycles when lubricated, which is sufficient for use in
SMORES-EP.
\subsection{Wheel and Tilt PaintPots in SMORES-EP}
Each of the four articulated joints in SMORES-EP is equipped with a
PaintPot, which provides absolute position encoding. The three
continuously-rotating faces (labeled TOP, LEFT, and RIGHT in
Fig.~\ref{fig:smores-ep}) have {\em Wheel PaintPots} with circular
tracks and wipers offset parallel to the axis of rotation, while the
central hinge has a {\em Tilt PaintPot} that covers \SI{180}{\degree} arc with
a wiper offset normal to the axis of rotation.
\subsubsection{Wheel PaintPots}
\label{sec:wheel-paintpot}
\begin{figure}[t]
  \centering
  \subfloat[]{\includegraphics[height=0.165\textwidth]{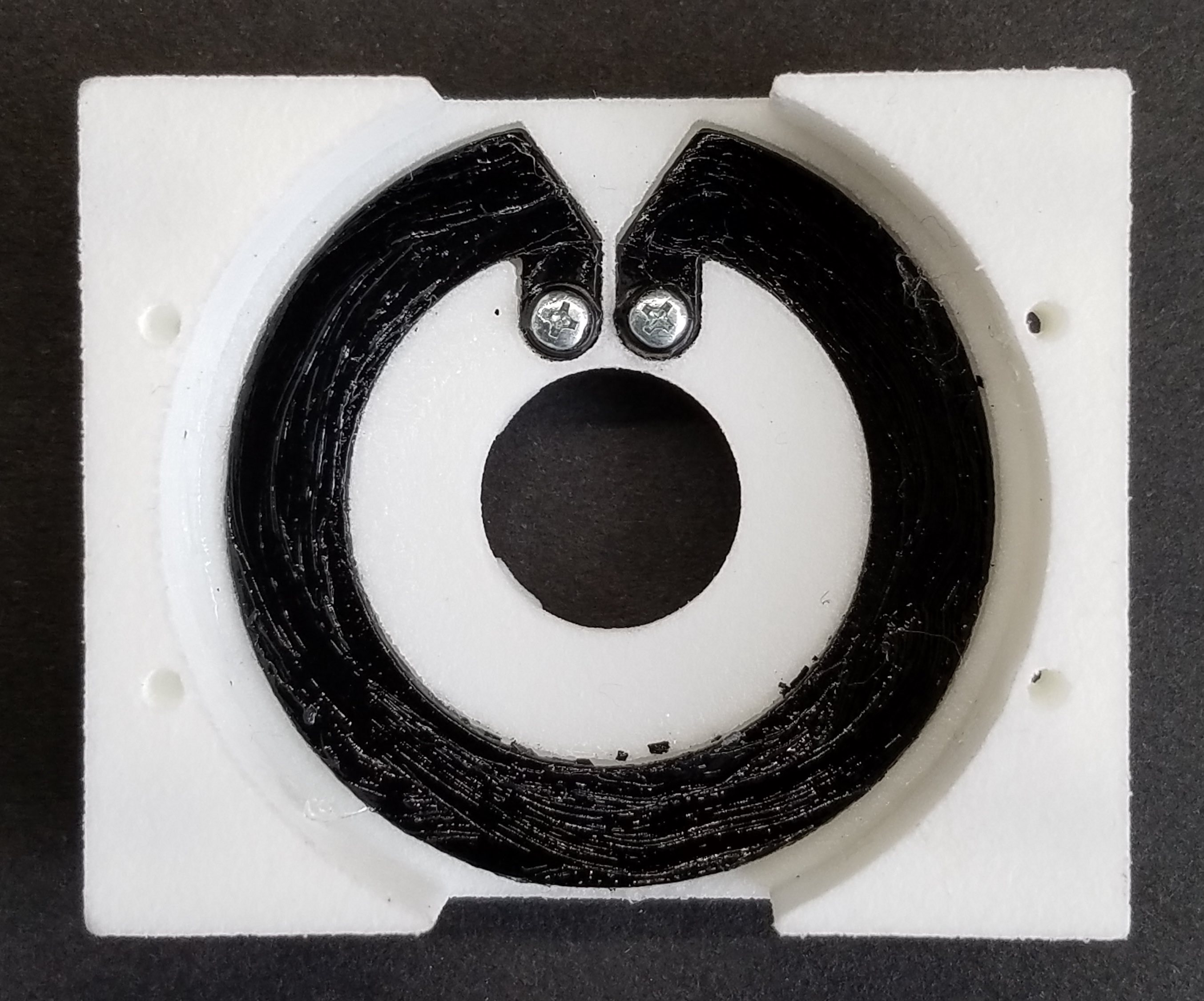}\label{fig:wheel-sensor}}
  \hfil
  \subfloat[]{\includegraphics[height=0.165\textwidth]{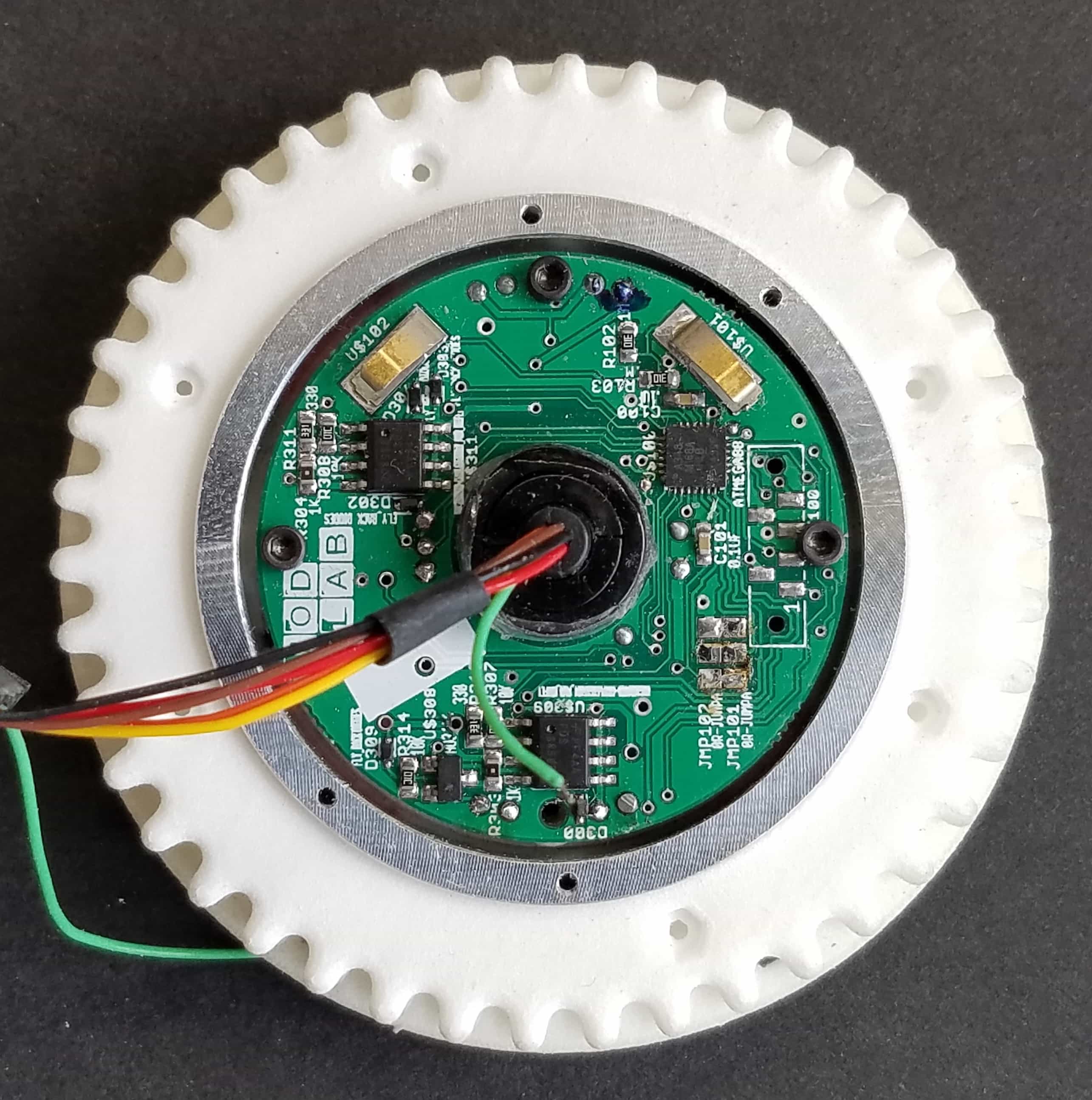}\label{fig:faceboard}}
  \caption{(a) A wheel PaintPot sensor is installed in a chassis. (b)
    Two wipers (Harwin S1791-42) are mounted on a circuit board at a
    \SI{50}{\degree} angle to one another fixed to the wheel.}
\end{figure}
The wheel PaintPots, shown in Fig.~\ref{fig:wheel-sensor}, have a
circular track and two wiper contacts, allowing continuous rotation
and providing position information over the full \SI{360}{\degree}
range of the left, right, and pan joints. The annular geometry allows
a slip ring to fit through the center.  Tabs on the track extend into
the center of the circle to provide space for the terminal
contacts. The V-shaped gap provides enough space for the wipers to
pass from one side of the track to the other without contacting both
simultaneously (which would cause a short circuit).  The two wipers
are mounted on a PCB above the track at a \SI{50}{\degree} angle to
one another (Fig.~\ref{fig:faceboard}); this configuration
ensures that at least one wiper contacts the track at all times.

Tracks are cut in batches in a laser cutter. To facilitate easy
mounting, a layer of double-sided adhesive is applied to the back of
the ABS sheet before cutting. After cutting, three coats of paint are
applied, and strips are allowed to dry for 24 hours.  Strips are
mounted in a mated groove in a 3D-printed chassis as shown in
Fig.~\ref{fig:wheel-sensor}. The chassis has a raised triangular
feature that mates with the gap in the strip, so that the wipers
remain at the same level as they pass through the gap
region. Zinc-coated screws are used for electrical terminals.  The
measured terminal-to-terminal resistance of wheel PaintPots ranges
from \SI{2}{\kilo\ohm} to \SI{20}{\kilo\ohm}, depending on the
thickness of the paint. Before use, a coat of petroleum-based grease
is applied to the track surface.
\subsubsection{Tilt PaintPots}
\label{sec:tilt-paintpot}
\begin{figure}[t]
  \centering
  \subfloat[]{\includegraphics[height=0.198\textwidth]{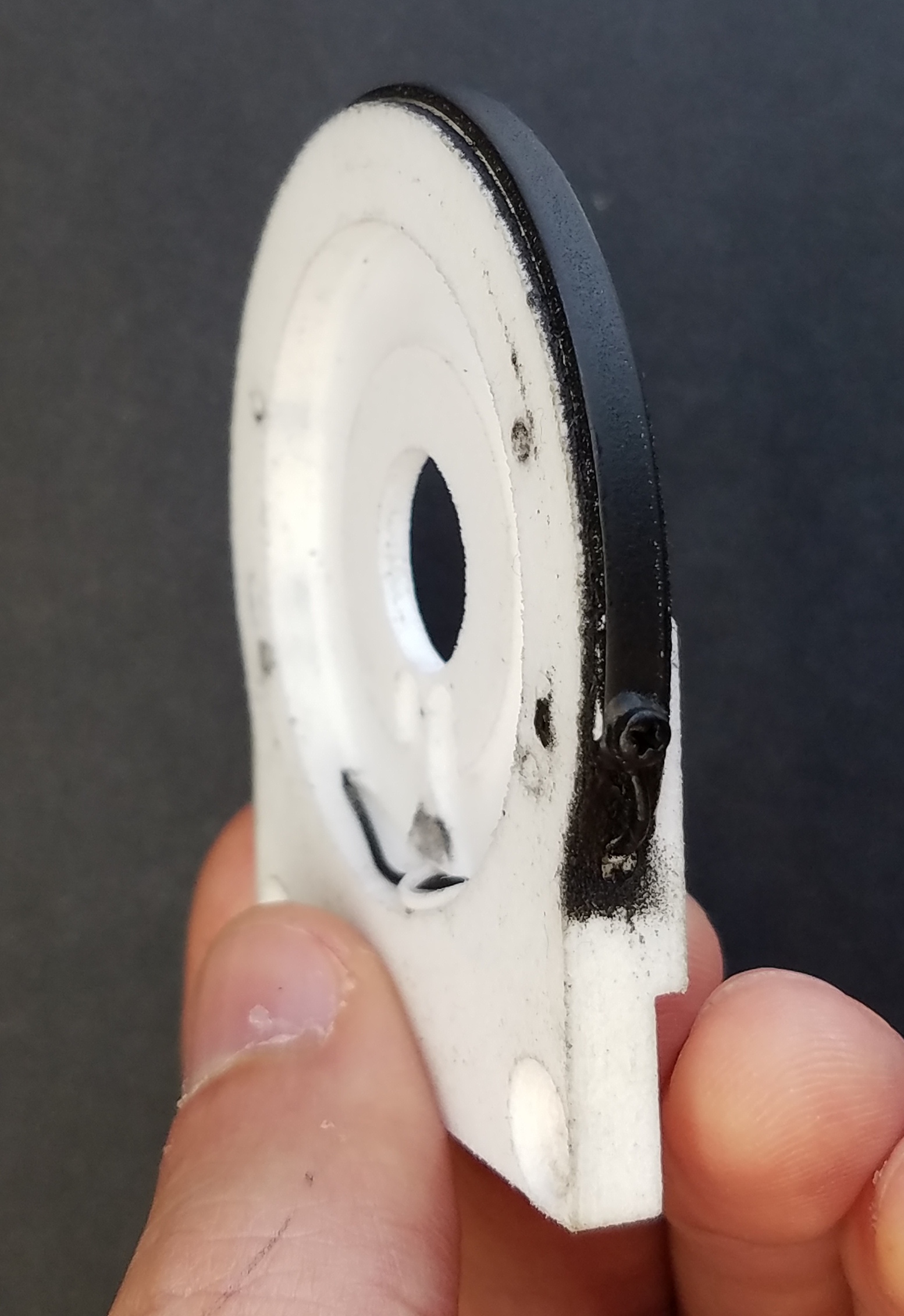}\label{fig:tilt-sensor}}
  \hfil
  \subfloat[]{\includegraphics[height=0.198\textwidth]{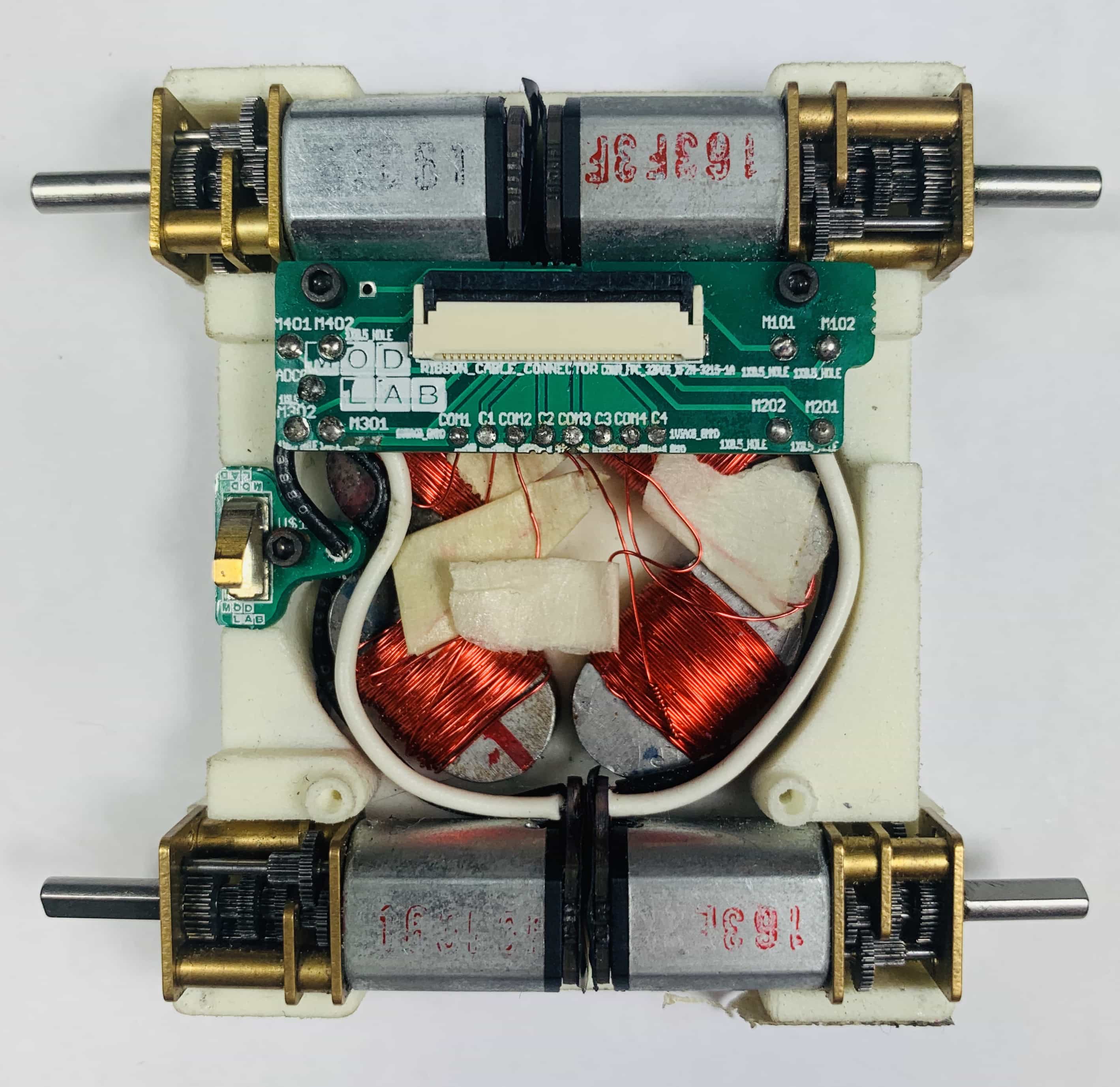}\label{fig:smores-base}}
  \caption{(a) A tilt PaintPot sensor is installed on a chassis. (b) A
  single wiper installed on the base of a SMORES-EP module contacts
  the track.}
\end{figure}
The tilt PaintPots, shown in Fig.~\ref{fig:tilt-sensor}, have tracks
with cylindrical curvature about their axis of rotation.  A single
wiper contacts the track and measures position through the full
\SI{180}{\degree} of motion of the tilt joint
(Fig.~\ref{fig:smores-base}). The track geometry of the tilt PaintPot
makes very efficient use of space inside the SMORES-EP module; to our
knowledge, no off-the-shelf potentiometers replicate this unusual
non-planar shape.

Tilt PaintPots have the same ABS/adhesive substrate as wheel
PaintPots, and similar screw contacts.  They are mounted to the 3D
printed chassis before painting, allowing them to be painted in their
final curved shape.  This is preferable to painting flat and then
bending: bending the paint after it has dried causes cracks to form,
increases the resistance (three orders of magnitude), and causes a
non-smooth variation of voltage along the length of the track. The
terminal-to-terminal resistances of the tilt PaintPots range from
\SI{3}{\kilo\ohm} to \SI{10}{\kilo\ohm}.
\subsection{Cost}
\label{sec:cost}
The PaintPots used in SMORES-EP are inexpensive. The wipers are
available from Digikey.com for \SI{0.35}[\$]{USD} in quantities of
100. A \SI{12}{oz} MG Chemicals Total Ground spray paint can be
purchased from Amazon.com for \SI{16}[\$]{USD}, and \SI{0.79}{mm} ABS
sheets can be purchased from McMaster.com for \SI{3.70}[\$]{USD} per
square foot. Based on these, materials for wheel PaintPots cost
\SI{1.05}[\$]{USD} and tilt PaintPots cost \SI{0.70}[\$]{USD}. In order to
build SMORES-EP modules with quality control testing~\cite{paintpot},
we yield about 75$\%$ of our wheel PaintPots and 90$\%$ of our tilt
PaintPots, making the effective materials costs \SI{1.40}[\$]{USD} and
\SI{0.78}[\$]{USD} respectively.

\section{Sensor Characterization}
\label{sec:characterization}

Potentiometers used as voltage dividers typically model the input
position as having linear relationship with the output voltage. Close
adherence to the linear model has to be achieved by ensuring that the
resistance between two points along the track is constant, which
requires uniform geometry, thickness, and material properties of the
track. This is difficult for PaintPots which are manually
spray-painted. In order to obtain accurate position control on all
DOFs of a SMORES-EP module, a calibration process is needed to
characterize the performance of the particular PaintPots
installed.

\begin{figure}[t]
  \centering
  \includegraphics[width=0.2\textwidth]{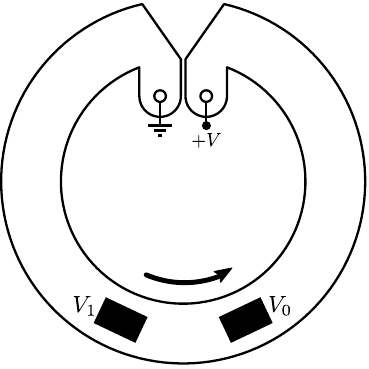}
  \caption{When the wheel position $\theta = \SI{0}{\radian}$, two wipers are contacting
    the track symmetrically to the middle location of it and $\theta$
    ranges from \SIrange{-\pi}{\pi}{\radian}.}
  \label{fig:wheel-zero}
\end{figure}

One terminal of a wheel PaintPot is connected with \SI{3.3}{\volt}
and the other terminal is connected with ground. Two wipers can
contact the track and report current voltage ($V_0$ and $V_1$) in the
form of two \SI{10}{\bit} analog-to-digital conversion values ranging
from \numrange{0}{1023}, and wheel position $\theta = \SI{0}{\radian}$ is shown in
Fig.~\ref{fig:wheel-zero} and the whole range of $\theta$ is from
\SIrange{-\pi}{\pi}{\radian}. When a wiper contacts on or around the V-shape gap,
the voltage value is not usable. This is the reason for having two
wipers, to enable sensing the full \SI{360}{\degree} range. So, $V_0$
should be ignored when $\theta$ is in the range from
\SIrange[parse-numbers=false]{\frac{2}{3}\pi}{\frac{5}{6}\pi}{\radian} and
$V_1$ should be ignored when $\theta$ is in the range from
\SIrange[parse-numbers=false]{-\frac{5}{6}\pi}{-\frac{2}{3}\pi}{\radian}.
\begin{figure}[t]
  \centering
  \includegraphics[width=0.4\textwidth]{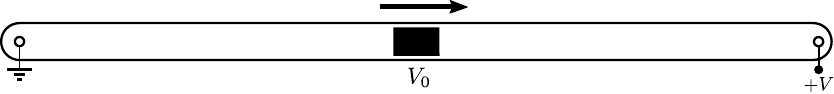}
  \caption{When the TILT DOF position $\theta = \SI{0}{\radian}$, the wiper
    is contacting the middle of the track and $\theta$ ranges from
    \SIrange[parse-numbers=false]{-\pi/2}{\pi/2}{\radian}.}
  \label{fig:tilt-zero}
\end{figure}

Similar to wheel PaintPots, the tilt PaintPot is also powered between
ground and \SI{3.3}{\volt} with one single wiper contacting the track
all the times. The voltage $V_0$ from the voltage divider goes through
a \SI{10}{\bit} analog-to-digital conversion (with range from
\numrange{0}{1023}). When the tilt position
$\theta = \SI{0}{\radian}$, the wiper is positioned in the middle of the
track as shown in Fig.~\ref{fig:tilt-zero}.

\begin{figure}[t]
  \centering
  \subfloat[]{\includegraphics[height=0.2\textwidth]{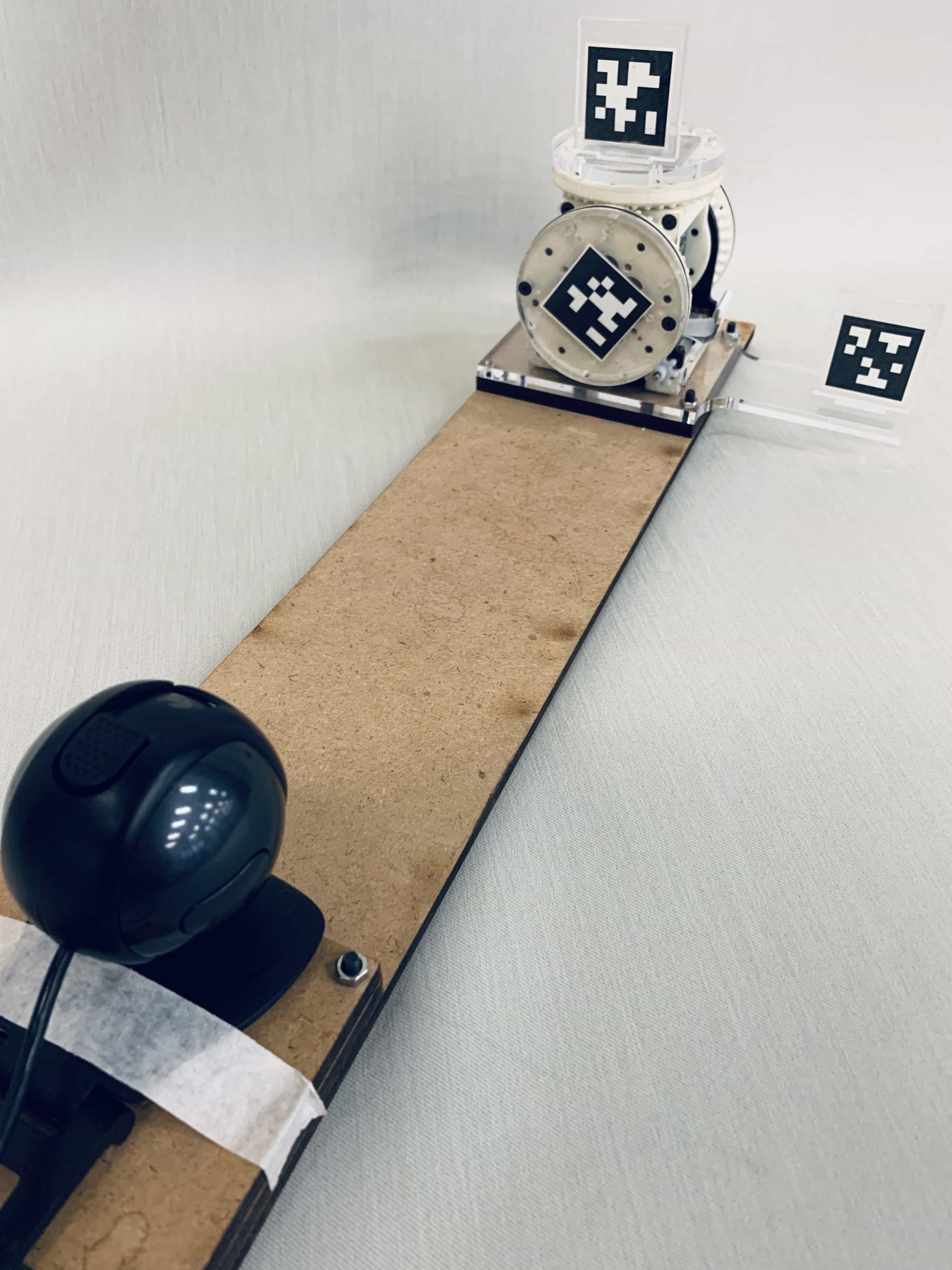}\label{fig:calibration-setup}}
  \hfil
  \subfloat[]{\includegraphics[height=0.2\textwidth]{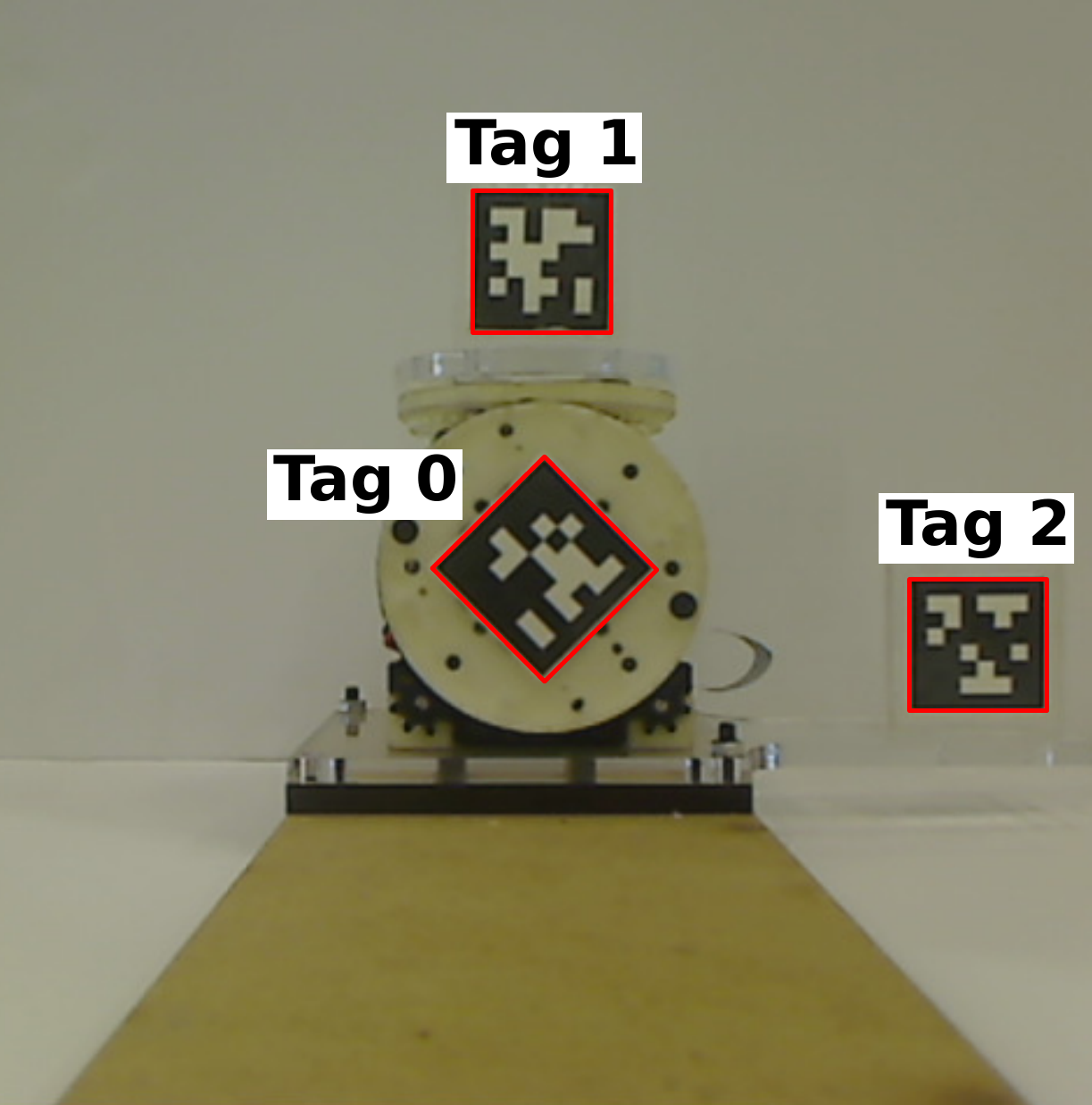}\label{fig:tag-image}}
  \caption{(a) Sensor characterization setup using AprilTags tracking
    approach. (b) Camera view of three tags in the characterization
    process.}
  \label{fig:sensor-calibration-setup}
\end{figure}

An automatic sensor calibration setup is developed based on AprilTags
tracking~\cite{apriltags} shown in
Fig.~\ref{fig:calibration-setup}. Three tags are used to track the
rigid bodies of a SMORES-EP module. Tag 2 is fixed to the base to be
the reference frame, Tag 1 is fixed to the TOP Face of a SMORES-EP
module for TILT DOF tracking, and Tag 0 can be fixed to LEFT Face,
RIGHT Face, or TOP Face for wheel DOF tracking
(Fig.~\ref{fig:tag-image}). During characterization, one DOF is moved
at a time through its entire range of motion (\SI{2\pi}{\radian} for
wheel DOF, \SI{\pi}{\radian} for TILT DOF) in both directions. The data,
including $\theta$ and reported voltage, are recorded at \SI{14}{Hz} (speed
limited by the AprilTag ROS package).

\begin{figure}[t]
  \centering
  \subfloat[]{\includegraphics[width=0.24\textwidth]{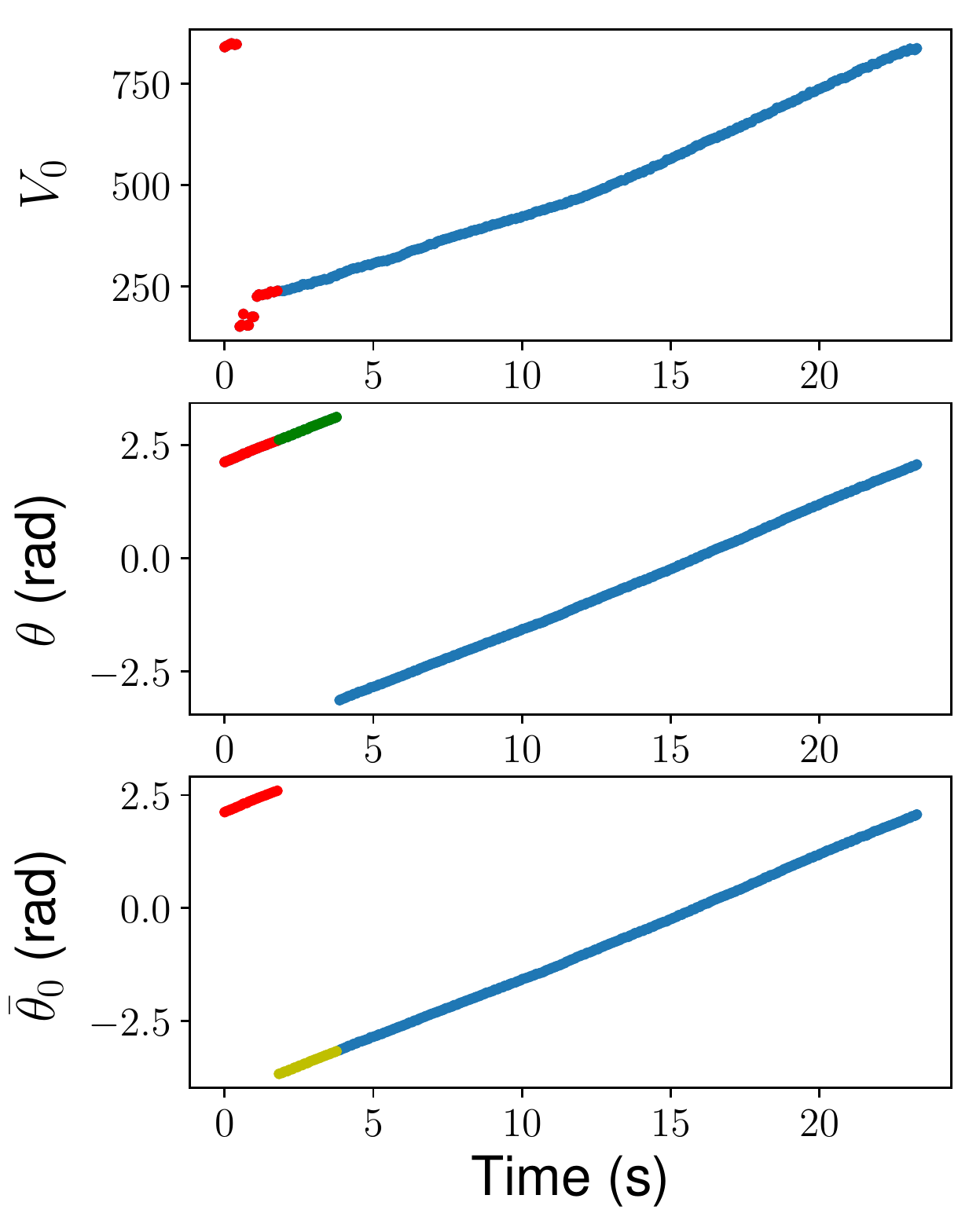}\label{fig:wiper0-data}}
  \hfil
  \subfloat[]{\includegraphics[width=0.24\textwidth]{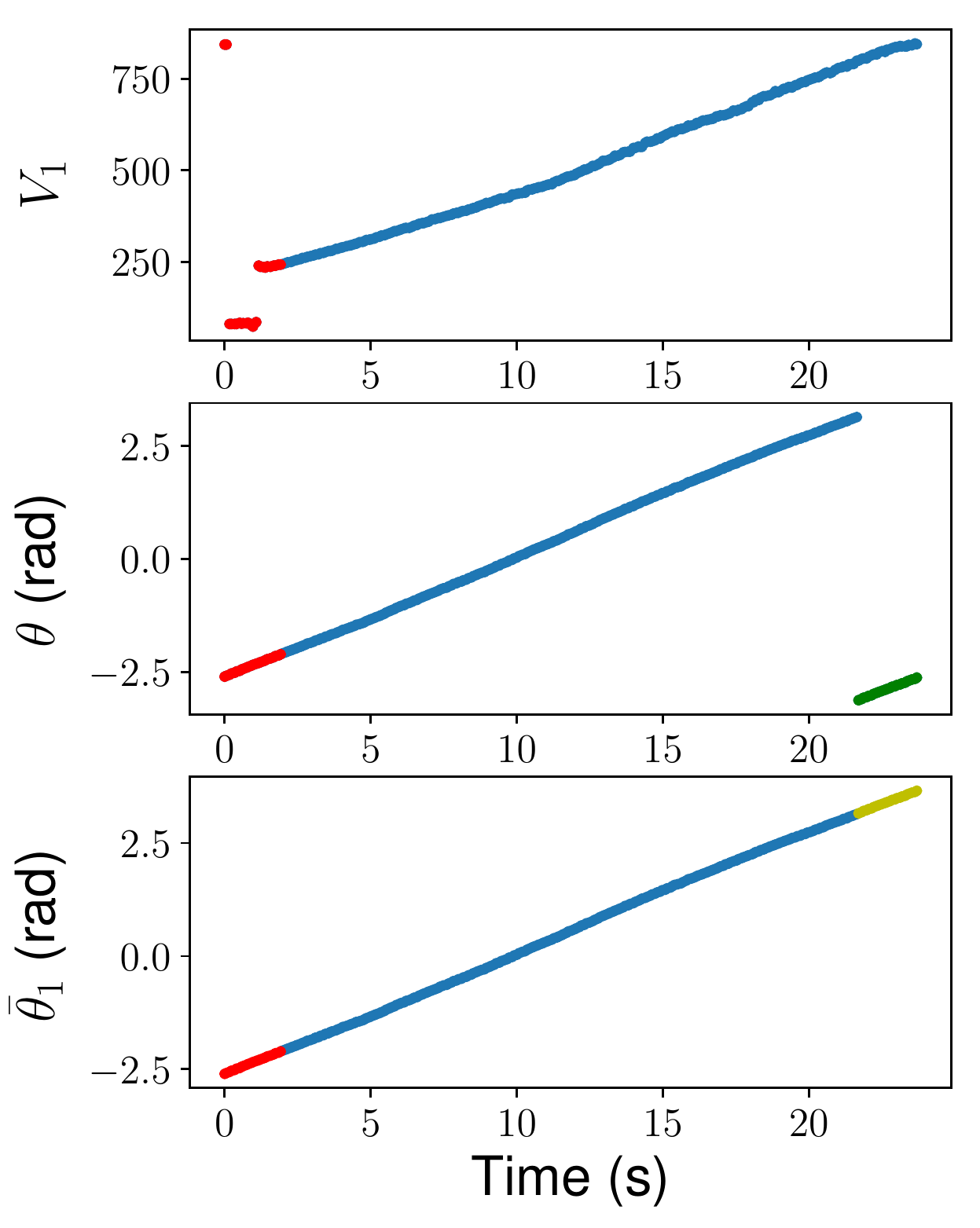}\label{fig:wiper1-data}}
  \caption{  (a) Wiper 0 data through the entire range of a wheel
    PaintPot. (b) Wiper 1 data through the entire range of a wheel PaintPot.}
\end{figure}

\begin{figure}[t]
  \centering
  \subfloat[]{\includegraphics[width=0.24\textwidth]{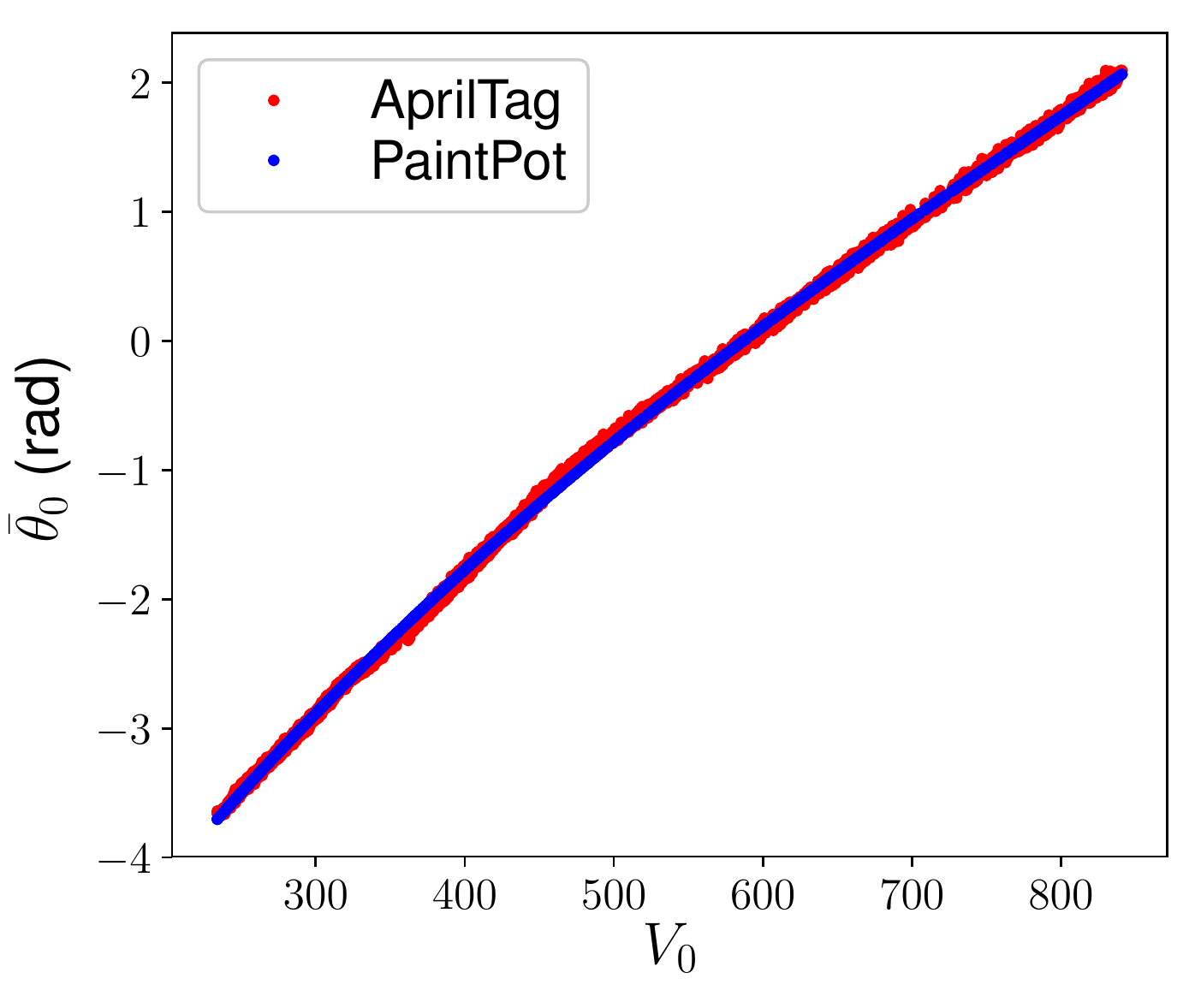}\label{fig:wiper0-measure}}
  \hfil
  \subfloat[]{\includegraphics[width=0.24\textwidth]{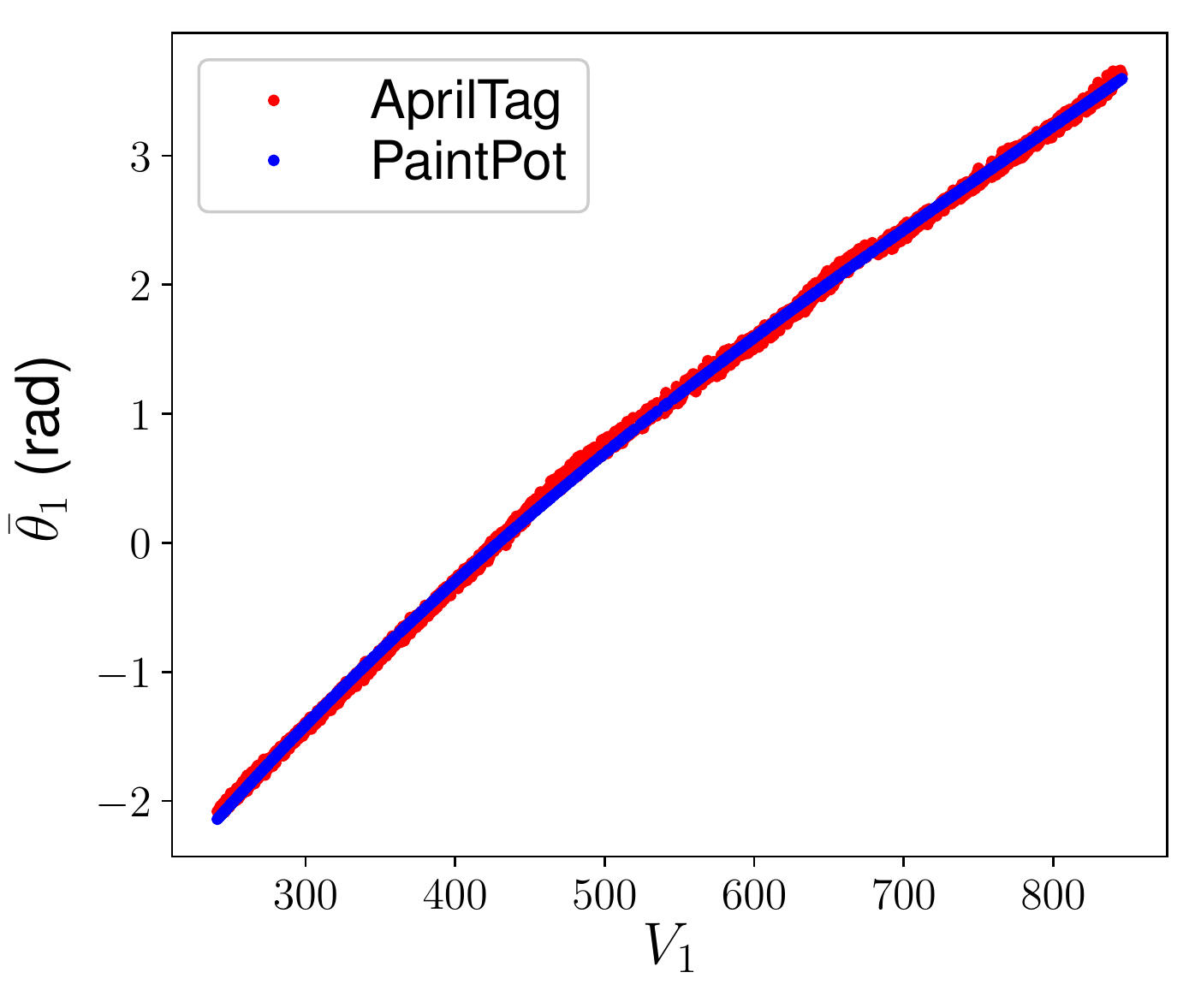}\label{fig:wiper1-measure}}
  \caption{
  Wheel PaintPot sensor characterization results: (a)
    $\bar{\theta}_0 = \bar{f}_0(V_0) = \num{5.0281e-9}V_0^3 - \num{1.2255e-5}V_0^2
    + \num{1.7856e-2}V_0-7.2750$ and (b)
    $\bar{\theta}_1 = \bar{f}_1(V_1) = \num{5.1596e-9}V_1^3 - \num{1.2409e-5}V_1^2
    + \num{1.7927e-2}V_1-5.8128$.}
\end{figure}

\begin{figure}[t]
  \centering
  \subfloat[]{\includegraphics[height=0.215\textwidth]{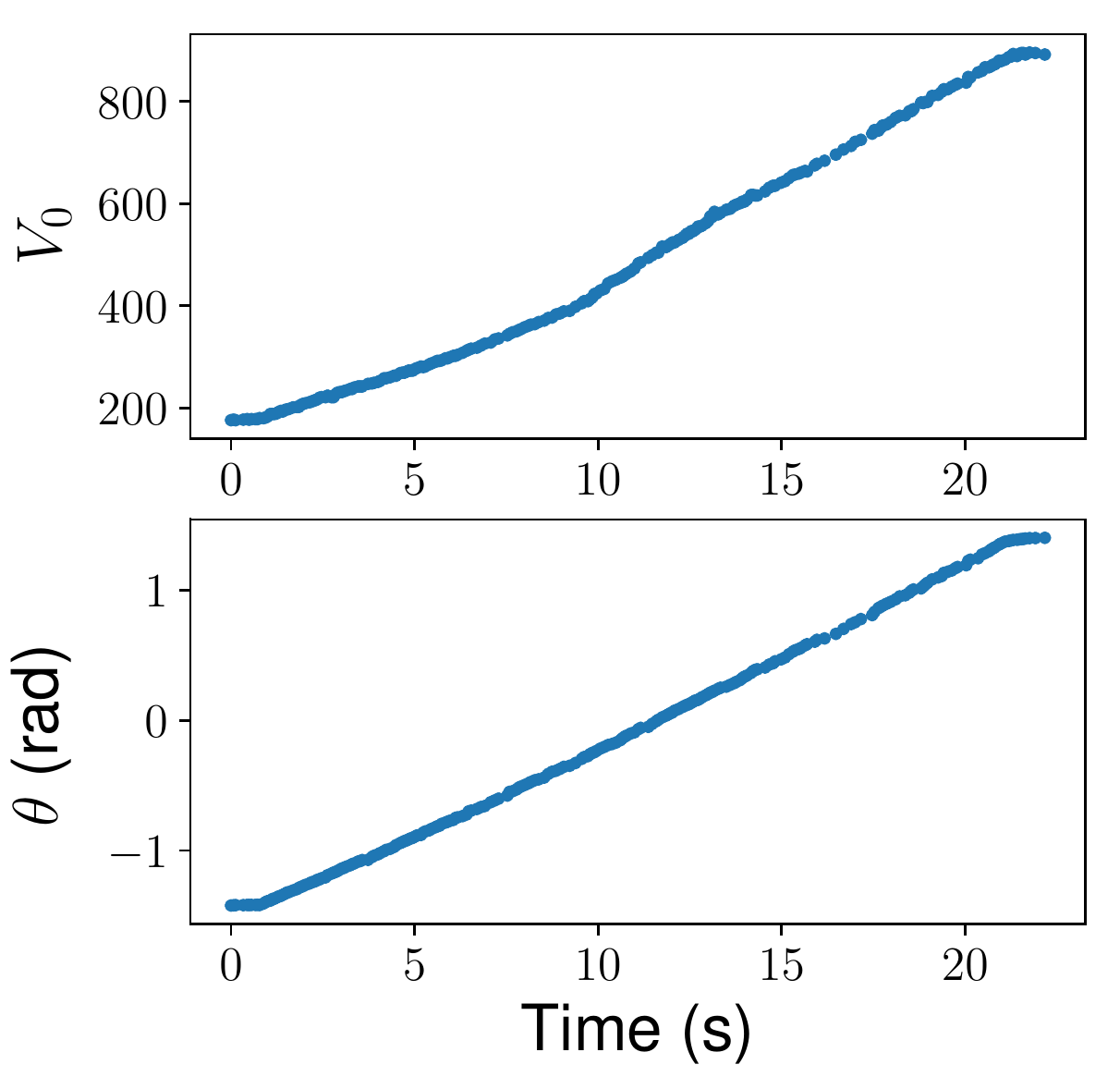}\label{fig:tilt-data}}
  \hfil
  \subfloat[]{\includegraphics[height=0.215\textwidth]{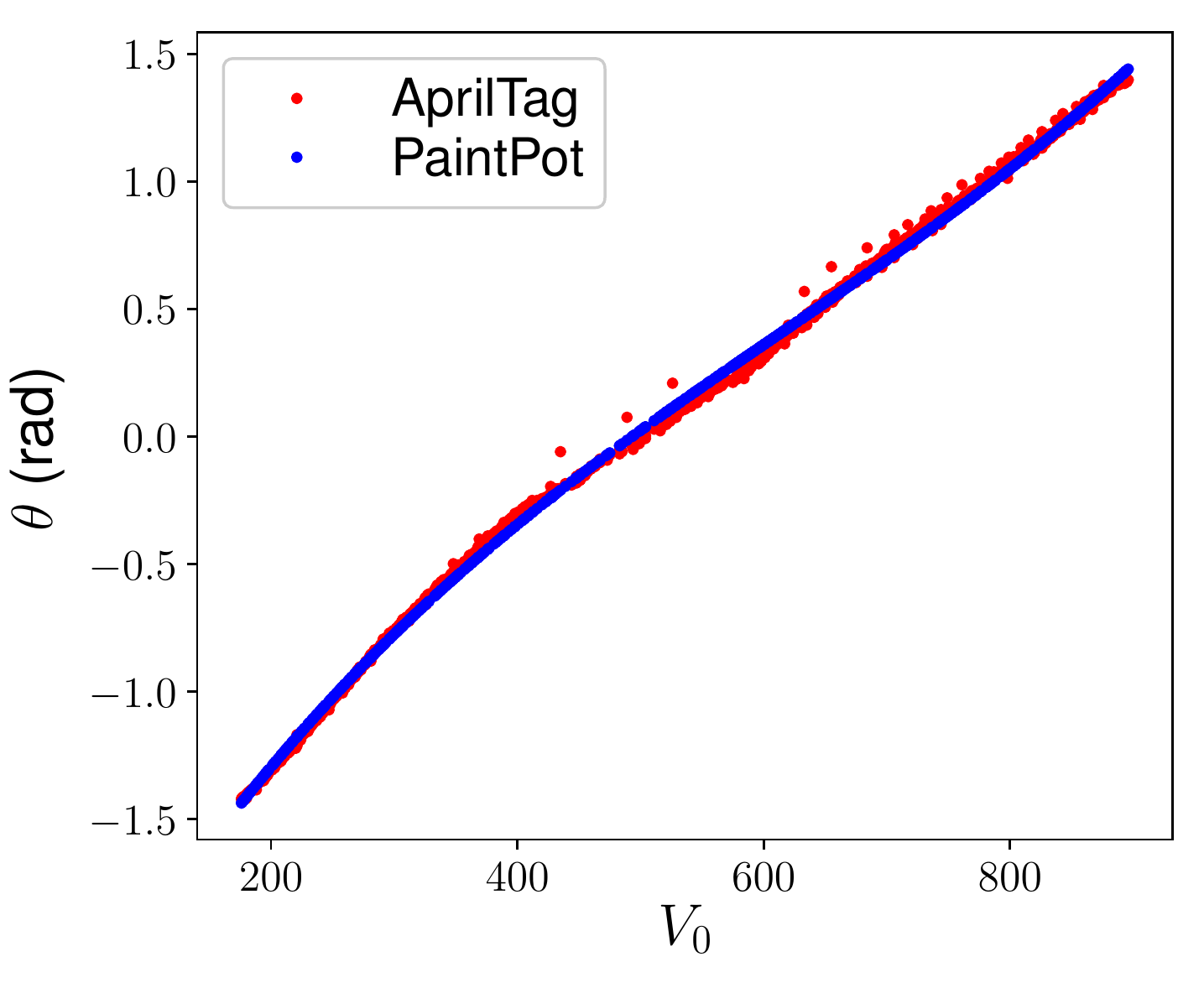}\label{fig:tilt-measure}}
  \caption{(a) Wiper data from \SIrange{-80}{80}{\degree} of a tilt
    PaintPot. (b) The characterization result $\theta = f(V_0) =
    \num{4.7517e-9}V_0^3 - \num{8.7608e-6}V_0^2 + \num{8.6756e-3}V_0 -
    2.7173$.}
\end{figure}

While the voltage data is not linear with DOF position, it is
monotonic (piece-wise monotonic for wheel DOFs). A third-order
polynomial provides a suitable model. For a wheel PaintPot, the data
from both wipers are shown in Fig.~\ref{fig:wiper0-data} and
Fig.~\ref{fig:wiper1-data} respectively. For wiper 0, the reported
voltage $V_0$ is not useful when DOF position $\theta$ is in the range from
\SIrange[parse-numbers=false]{\frac{2}{3}\pi}{\frac{5}{6}\pi}{\radian}
(shown in red). Due to the gap of wheel PaintPots, $\theta = f_0(V_0)$ is a
piece-wise function which can be converted into a continuous function
$\bar{\theta}_0 = \bar{f}_0(V_0)$ by shifting the segment ranging from
\SIrange[parse-numbers=false]{\frac{5}{6}\pi}{\pi}{\radian} (shown in
green) by \SI{2\pi}{\radian} downward (shown in yellow). Similarly, for
wiper 1, the segment when $\theta$ is in the range from
\SIrange[parse-numbers=false]{-\frac{5}{6}\pi}{-\frac{2}{3}\pi}{\radian}
(shown in red) is meant to be trimmed, and the piece-wise function
$\theta = f_1(V_1)$ is converted into a continuous function
$\bar{\theta}_1 = \bar{f}_1(V_1)$ by shifting the segment ranging from
\SIrange[parse-numbers=false]{-\pi}{-\frac{5}{6}\pi}{\radian} (shown in
green) by \SI{2\pi}{\radian} upward (shown in yellow). After taking
\SI{50}{s} data in both directions (showing little hysteresis),
$\bar{f}_0(V_0)$ and $\bar{f}_1(V_1)$ are shown in
Fig.~\ref{fig:wiper0-measure} and Fig.~\ref{fig:wiper1-measure}
respectively. An example run from a tilt PaintPot is shown in
Fig.~\ref{fig:tilt-data} and the characterization result
$\theta= f(V_0)$ is shown in Fig.~\ref{fig:tilt-measure}.

\section{Position Estimation}
\label{sec:estimate}

\subsection{Transition Model}
\label{sec:transition}

\begin{figure}[t]
  \centering
  \includegraphics[width=0.4\textwidth]{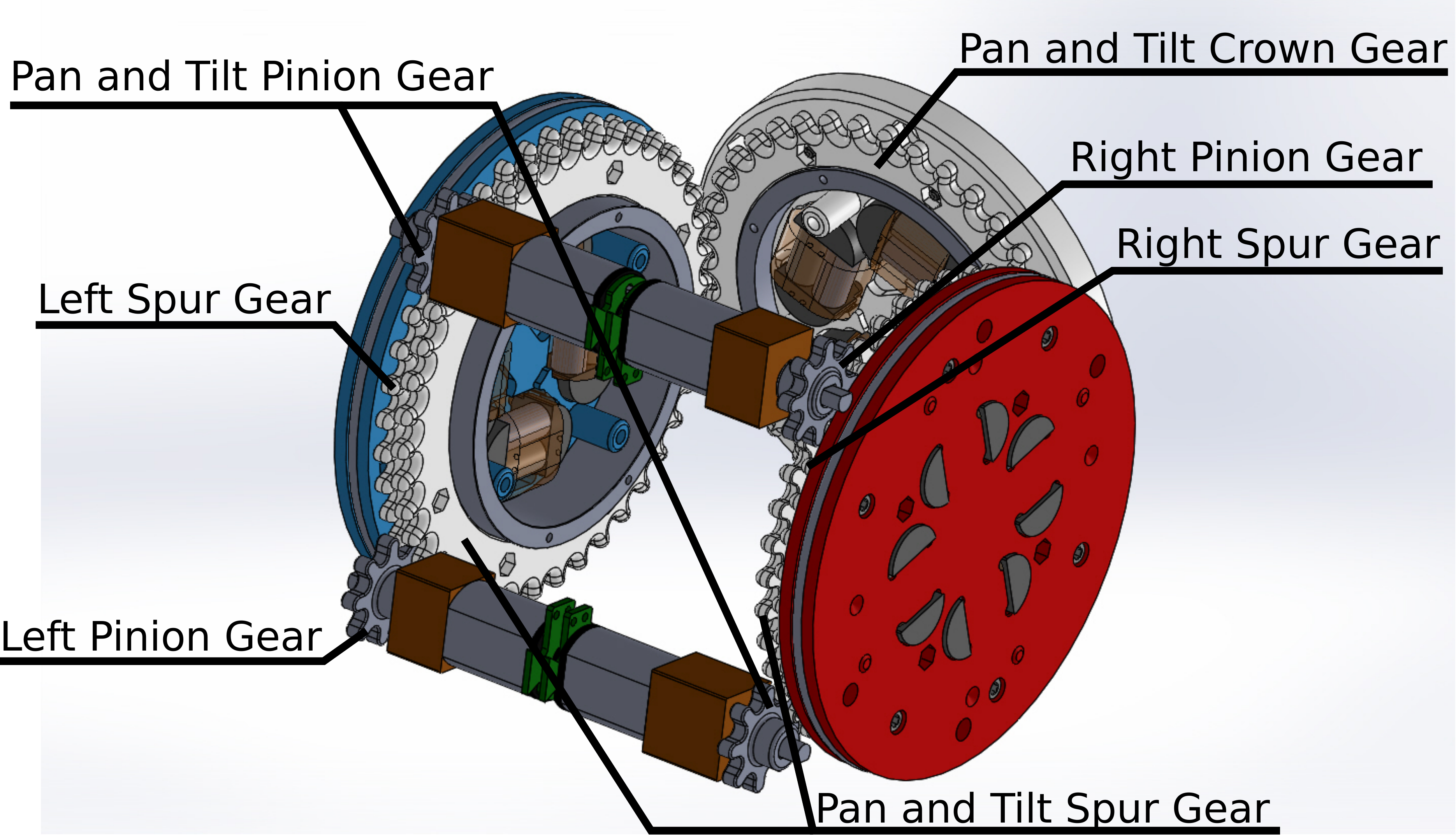}
  \caption{Four DOFs are driven by four actuators thorough a gear train.}
  \label{fig:smores-driving}
\end{figure}

Four DC motors are used to drive four DOFs (LEFT DOF, RIGHT DOF, PAN
DOF, and TILT DOF) with a geared drive train shown in
Fig.~\ref{fig:smores-driving}. The drive has pinion gears driving four
identical spur gears. Two outer spur gears are attached to the left
wheel and right wheel respectively for the LEFT DOF and RIGHT DOF. The
crown gear is coupled to the two inner spur gears. When these two
inner gears spin in the opposite direction, the top wheel attached to
the crown gear rotates which is the PAN DOF. TILT DOF rotates when
these two inner spur gears spin in the same direction. The
transmission ratio for each DOF is determined by the gear train, then
a linear relationship between the DOF velocity and the motor angular
velocity can be obtained
\begin{equation}
  \label{eq:transmission}
  \dot{\theta} = k\omega + n
\end{equation}
in which $k$ is the transmission ratio of the DOF, $\omega$ is the angular
velocity of the driving motor(s), $n\sim N(0, Q)$ is the additive
Gaussian white noise, and $\dot{\theta}$ is the angular velocity of the
DOF. The transition model of a DOF in discrete time can be derived by
one-step Euler integration:
\begin{equation}
  \label{eq:transition}
  \begin{aligned}
    \theta_t &= \theta_{t-1} + \dot{\theta}_{t-1} \delta t + n_{t-1}\delta t\\
    &= \theta_{t-1} + k\omega_{t-1}\delta t + n_{t-1}\delta t\\
    &= \theta_{t-1} + G\omega_{t-1} + Un_{t-1}
  \end{aligned}
\end{equation}
in which $\delta t$ is the finite time interval. Let $\theta$ be the state
$x$ and $\omega$ be the system input $u$, then the transition model is
\begin{equation}
  \label{eq:transition-model}
  x_t = x_{t-1} + Gu_{t-1} + Un_{t-1}
\end{equation}

\subsection{Observation Model}
\label{sec:observation}

In Section~\ref{sec:characterization}, a PaintPot can be characterized
with a nonlinear model --- a third-order polynomial $\theta = f(V)$. Here
$V$ is the measurement, namely the reported voltage(s) from the
wiper(s). Based on this, a nonlinear observation model $z = h(x)$
where $z$ is the measurement and $x$ is the state can be derived. We
can linearize $z = h(x)$ about the current mean and variance of the
DOF position $x$ to apply an extended Kalman filter (EKF). However,
this places some computational burden on the SMORES-EP
microcontroller. Here, we present a simple approach that can generate
a linear observation model directly without linearization of the
original nonlinear observation model about the current mean and
variance of the state to overcome this nonlinearity. With this new
approach, we just need to check if features are available using a
simple rule and compute the new converted measurements by evaluating
polynomials obtained from sensor characterization
(Section~\ref{sec:characterization}).

\subsubsection{Wheel PaintPots}
\label{sec:wheel-paintpot-observe}

For the wheel PaintPot sensor, the observation model is a piece-wise
function due to the geometry of the sensor:
\begin{enumerate}
\item When $\theta\in (\SI{-\pi}{\radian}, \SI[parse-numbers=false]{-\frac{5}{6}\pi}{\radian})\cup (\SI[parse-numbers=false]{-\frac{2}{3}\pi}{\radian}, \SI[parse-numbers=false]{\frac{2}{3}\pi}{\radian})\cup
  (\SI[parse-numbers=false]{\frac{5}{6}\pi}{\radian}, \SI{\pi}{\radian})$, there are two valid measurements which are the
  reported voltages $V_0$ and $V_1$ from both wipers;
\item When
  $\theta\in \left[\SI[parse-numbers=false]{-\frac{5}{6}\pi}{\radian},
    \SI[parse-numbers=false]{-\frac{2}{3}\pi}{\radian}\right]$, only
  $V_1$ is valid;
\item When
  $\theta\in \left[\SI[parse-numbers=false]{\frac{2}{3}\pi}{\radian},
    \SI[parse-numbers=false]{\frac{5}{6}\pi}{\radian}\right]$, only
  $V_0$ is valid.
\end{enumerate}

To accommodate the \SI{2\pi}{\radian} shift to obtain a continuous
function for sensor characterization
(Section~\ref{sec:characterization}), the observation can be modeled
as
\begin{equation}
  \label{eq:observation-wheel-raw}
  \begin{aligned}
    z =
    \begin{dcases}
      \left[
        \begin{matrix}
          V_0\\V_1
        \end{matrix}
      \right] =
      \left[
        \begin{array}{c}
          \bar{f}_0^{-1}(x)\\
          \bar{f}_1^{-1}(x+2\pi)
        \end{array}
      \right],& x<\SI[parse-numbers=false]{-\frac{5}{6}\pi}{\radian}\\
      V_1 = \bar{f}^{-1}_1(x),&\SI[parse-numbers=false]{-\frac{5}{6}\pi}{\radian} \le x \le \SI[parse-numbers=false]{-\frac{2}{3}\pi}{\radian}\\
      \left[
        \begin{matrix}
          V_0\\ V_1
        \end{matrix}
      \right] =
      \left[
      \begin{array}{c}
        \bar{f}_0^{-1}(x)\\
        \bar{f}_1^{-1}(x)
      \end{array}\right],& \SI[parse-numbers=false]{-\frac{2}{3}\pi}{\radian} < x < \SI[parse-numbers=false]{\frac{2}{3}\pi}{\radian}\\
    V_0 = \bar{f}_0^{-1}(x),&\SI[parse-numbers=false]{\frac{2}{3}\pi}{\radian} \le x \le \SI[parse-numbers=false]{\frac{5}{6}\pi}{\radian}\\
    \left[
      \begin{matrix}
        V_0\\ V_1
      \end{matrix}
    \right] =
    \left[
      \begin{array}{c}
        \bar{f}_0^{-1}(x-2\pi)\\
        \bar{f}_1^{-1}(x)
      \end{array}
    \right],& x > \SI[parse-numbers=false]{\frac{5}{6}\pi}{\radian}
  \end{dcases}
  \end{aligned}
\end{equation}
in which $x$ is the DOF position $\theta$.

In order to avoid linearizing this piece-wise observation model, we change the
measurement to be the reported states rather than the two reported
voltages. During the motion, there is at least one feature available
for tracking, namely at least one wiper is contacting the wheel
PaintPot at any time. Let $z^i$ be the measurement from the $i$th
feature, then the measurement model with additive Gaussian white noise
is simplified as
\begin{equation}
  \label{eq:observation-wheel}
  z^i = x^i  + v^i = h^i(x, v^i)\quad i=0,1
\end{equation}
in which $v^i\sim N(0, R^i)$, and $x^i$ is
the reported state from $i$th feature determined by state $x$ in the
following way:
\begin{subequations}
  \begin{equation}
    \label{eq:observation-0-case}
    x^0 =
    \begin{dcases}
      x&x\le \frac{5}{6}\pi\\
      x-2\pi&x> \frac{5}{6}\pi
    \end{dcases}
  \end{equation}
  \begin{equation}
    \label{eq:observation-1-case}
    x^1 =
    \begin{dcases}
      x+2\pi&x< -\frac{5}{6}\pi\\
      x&x\ge -\frac{5}{6}\pi
    \end{dcases}
  \end{equation}
\end{subequations}

\noindent Here, the measurement model is linear and the predicted
measurement can be computed easily. The actual measurement for the
$i$th feature can be obtained by evaluating $\bar{f}_i(V_i)$ derived
from the sensor characterization process. Recall that when
$x\in \left[\SI[parse-numbers=false]{\frac{2}{3}\pi}{\radian},
  \SI[parse-numbers=false]{\frac{5}{6}\pi}{\radian}\right]$, $V_0$ is
not valid meaning this feature is not available. Otherwise the actual
measurement from this feature is simply $\bar{f}_0(V_0)$ if
$x\notin \left[\SI[parse-numbers=false]{\frac{2}{3}\pi}{\radian},
  \SI[parse-numbers=false]{\frac{5}{6}\pi}{\radian}\right]$. And the
valid range of $V_0$ is from
$V^0_{\min} = \bar{f}_0^{-1}(\frac{5}{6}\pi-2\pi)$ (because the segment
from \SI[parse-numbers=false]{\frac{5}{6}\pi}{\radian} to
\SI{\pi}{\radian} is shifted downward by \SI{2\pi}{\radian}) to
$V^0_{\max} = \bar{f}_0^{-1}(\frac{2}{3}\pi)$.  Similar procedures can
be applied to $V_1$, this feature is $\bar{f}_1(V_1)$ if
$x\notin \left[\SI[parse-numbers=false]{-\frac{5}{6}\pi}{\radian},
  \SI[parse-numbers=false]{-\frac{2}{3}\pi}{\radian}\right]$, otherwise
this feature is not available. The valid range of $V_1$ is from
$V_{\min}^1 = \bar{f}_1^{-1}(-\frac{2}{3}\pi)$ to
$V_{\max}^{1} = \bar{f}_1^{-1}(-\frac{5}{6}\pi + 2\pi)$.

\subsubsection{Tilt PaintPots}
\label{sec:tilt-paintpot-observe}

The observation model for tilt PaintPots is
straightforward which is
\begin{equation}
  \label{eq:observation-tilt-raw}
  z = V_0 = f^{-1}(x)
\end{equation}
and it is a nonlinear function. Similarly, in order to avoid
linearizing this observation model, we change the measurement to be
the reported state and there is only one feature for tracking. Then
the observation model with additive Gaussian white noise is simplified
as
\begin{equation}
  \label{eq:observation-tilt}
  z = x + v = h(x, v)
\end{equation}
in which $v\sim N(0, R)$ and the model is linear. The feature should
always be available and the actual measurement for this feature is
obtained by evaluating $f(V_0)$.

\subsection{Kalman Filter}
\label{sec:kalman}

With the transition model and the new form of observation model, a Kalman
filter framework can be applied for state estimation.

\subsubsection{Kalman Filter for Wheels}
\label{sec:kalman-wheel}

The initial state of a wheel DOF can be derived by any available
feature. First check $V_0$, and if it is inside the valid range from
$V_{\min}^0$ to $V_{\max}^0$, compute the initial state
$x_0 = \bar{f}_0(V_0)$, and shift $x_0$ if necessary. That is if
$x_0 < -\pi$, let $x_0$ be $x_0 + 2\pi$. If
$V_0\notin (V_{\min}^0, V_{\max}^0)$, then $x_0 = \bar{f}_1(V_1)$ because
the wiper 1 must contact valid range of the track at this time, and
similarly shift $x_0$ if necessary, namely if $x_0 > \pi$, let $x_0$ be
$x_0 - 2\pi$. The prior state can be represented as a Gaussian
distribution $p(x_0)\sim N(\mu_0, \Sigma_0)$ where $\mu_0 = x_0$ and
$\Sigma_0$ is initialized to an arbitrarily small value.

With Eq.~\eqref{eq:transition-model}, the prediction step is
\begin{subequations}
  \begin{equation}
    \label{eq:predict-state}
    \bar{\mu}_t = \mu_{t-1} + Gu_t
  \end{equation}
  \begin{equation}
    \label{eq:predict-variance}
    \overline{\Sigma}_t = \Sigma_{t-1} + U^2Q
  \end{equation}
\end{subequations}
With Eq.~\eqref{eq:observation-wheel},
Eq.~\eqref{eq:observation-0-case}, and
Eq.~\eqref{eq:observation-1-case}, the predicted measurement
$\bar{z}_t$ that is related to $x^i_t$ (the reported state from $i$th
feature with $i=0,1$) can be computed. The analog-to-digital value
$V_i$ from the $i$th feature at time $t$ is used to compute the actual
measurement $z^i$. If $V_i \in (V^i_{\min}, V^i_{\max})$,
$z^i_t = \bar{f}_i(V_i)$. Otherwise, this feature is not available. If
both features are available, then
$z_t = \left[z^0_t, z^1_t\right]^\intercal$,
$C_t = \left[1, 1\right]^\intercal$, and the Kalman gain is
\begin{equation}
  \label{eq:kalman-gain-both}
  K_t = \overline{\Sigma}_t C_t^\intercal(C_t\overline{\Sigma}_tC_t^\intercal + R)^{-1}
\end{equation}
in which $R = \mathrm{diag}(R^0, R^1)$.
If there is only one feature (e.g. the $i$th feature) available, then $z_t =
z^i_t$, $C_t = 1$, and the Kalman gain is
\begin{equation}
  \label{eq:kalman-gain-one}
  K_t = \overline{\Sigma}_tC_t(C_t^2\overline{\Sigma}_t + R^i)^{-1}
\end{equation}
The state is then updated:
\begin{subequations}
  \begin{equation}
    \label{eq:state-update-wheel}
    \mu_t = \bar{\mu}_t + K_t(z_t - \bar{z}_t)
  \end{equation}
  \begin{equation}
    \label{eq:variance-update-wheel}
    \Sigma_t = \overline{\Sigma}_t - K_tC_t\overline{\Sigma}_t
  \end{equation}
\end{subequations}
The estimated position for this wheel DOF at time $t$ is $\mu_t$.

\subsubsection{Kalman Filter for Tilt}
\label{sec:kalman-tilt}

The initial state for a TILT DOF can be derived by evaluating
$f(V_0)$. The prior state can be represented as a Gaussian
distribution $p(x_0)\sim N(\mu_0, \Sigma_0)$ where $\mu_0 = x_0$ and
$\Sigma_0$ is initialized with some small value. The prediction step is in
the same form with wheel DOFs (Eq.~\eqref{eq:predict-state} and
Eq.~\eqref{eq:predict-variance}). The predicted measurement can be
computed from Eq.~\eqref{eq:observation-tilt} which is simply
$\bar{\mu}_t$. The current actual measurement is computed by evaluating
$z_t = f(V_0)$ where $V_0$ is the current reported voltage. Then the
Kalman gain is simply
\begin{equation}
  \label{eq:kalman-gain-tilt}
  K_t = \overline{\Sigma}_t(\overline{\Sigma}_t + R)^{-1}
\end{equation}
and the state is updated in the following:
\begin{subequations}
  \begin{equation}
    \label{eq:state-update-tilt}
    \mu_t = \bar{\mu}_t + K_t(z_t - \bar{\mu}_t)
  \end{equation}
  \begin{equation}
    \label{eq:variance-update-tilt}
    \Sigma_t = \overline{\Sigma}_t - K_t\overline{\Sigma}_t
  \end{equation}
\end{subequations}
And the estimated position for TILT DOF at time $t$ is $\mu_t$.

\section{Experiment}
\label{sec:experiment}

The PaintPot sensors, including wheel PaintPots and tilt PaintPots,
are installed in SMORES-EP modules for position sensing. Currently, 25
SMORES-EP modules have been assembled. In the experiments, both a
wheel PaintPot and a tilt PaintPot are characterized first and then
the newly developed Kalman filters are implemented to show the
effectiveness of our low-cost position sensing solution.

\begin{figure}[b]
  \centering
  \subfloat[]{\includegraphics[width=0.24\textwidth]{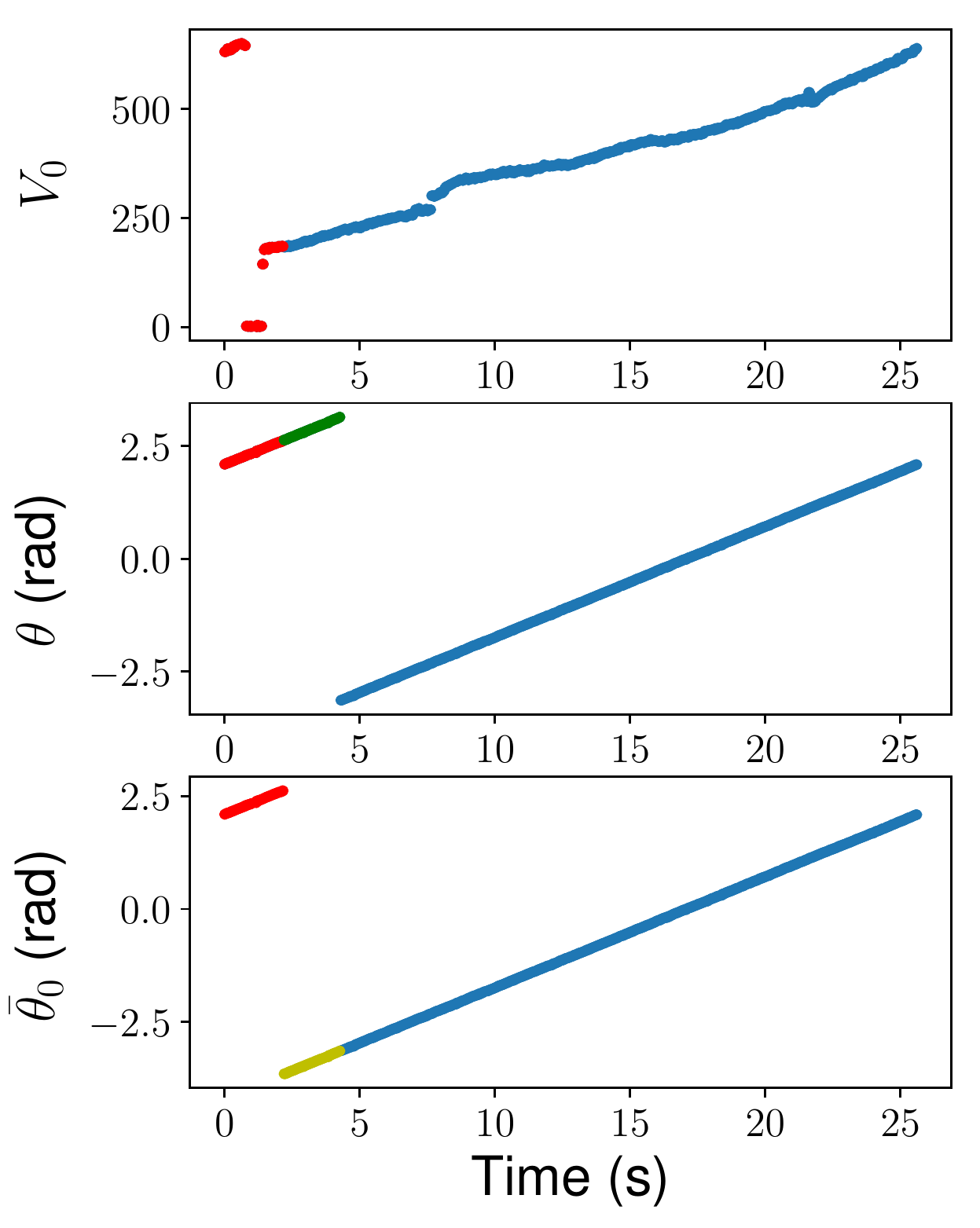}\label{fig:wiper0-experiment-data}}
  \hfil
  \subfloat[]{\includegraphics[width=0.24\textwidth]{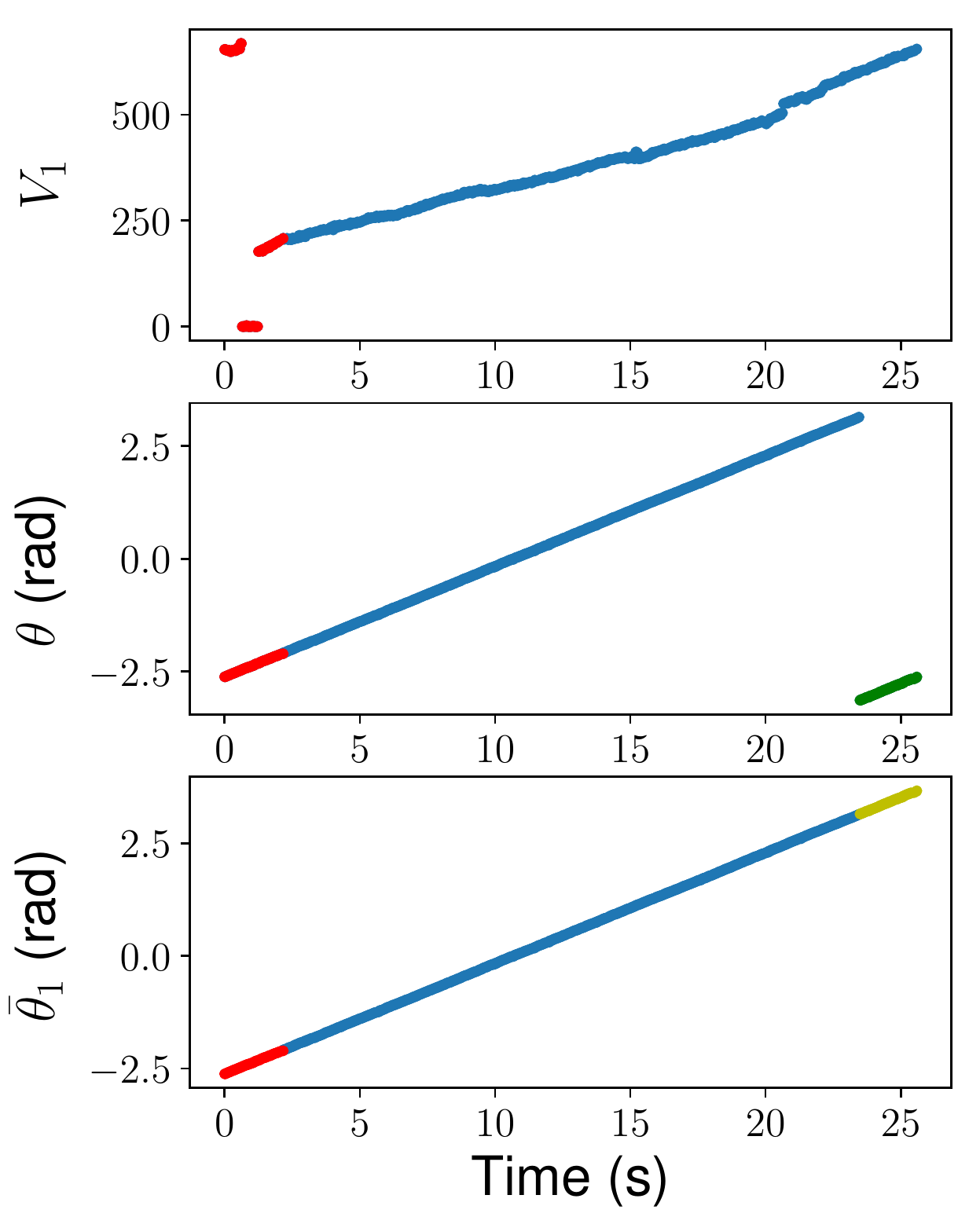}\label{fig:wiper1-experiment-data}}
  \caption{(a) Wiper 0 data through the entire range of a wheel
    PaintPot. (b) Wiper 1 data through the entire range of a wheel
    PaintPot.}
\end{figure}

\begin{figure}[t]
  \centering
  \subfloat[]{\includegraphics[width=0.24\textwidth]{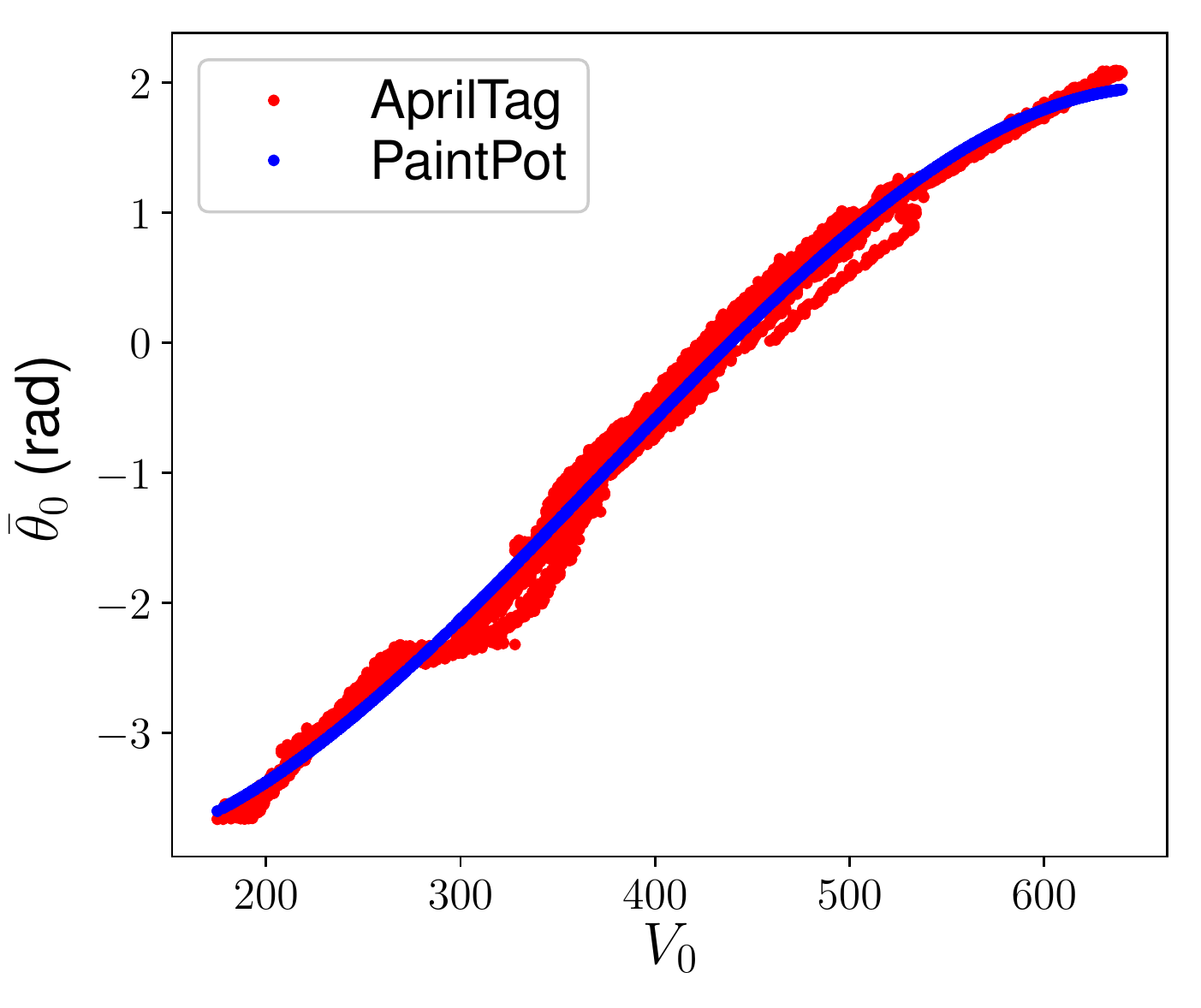}\label{fig:wiper0-experiment-measure}}
  \hfil
  \subfloat[]{\includegraphics[width=0.24\textwidth]{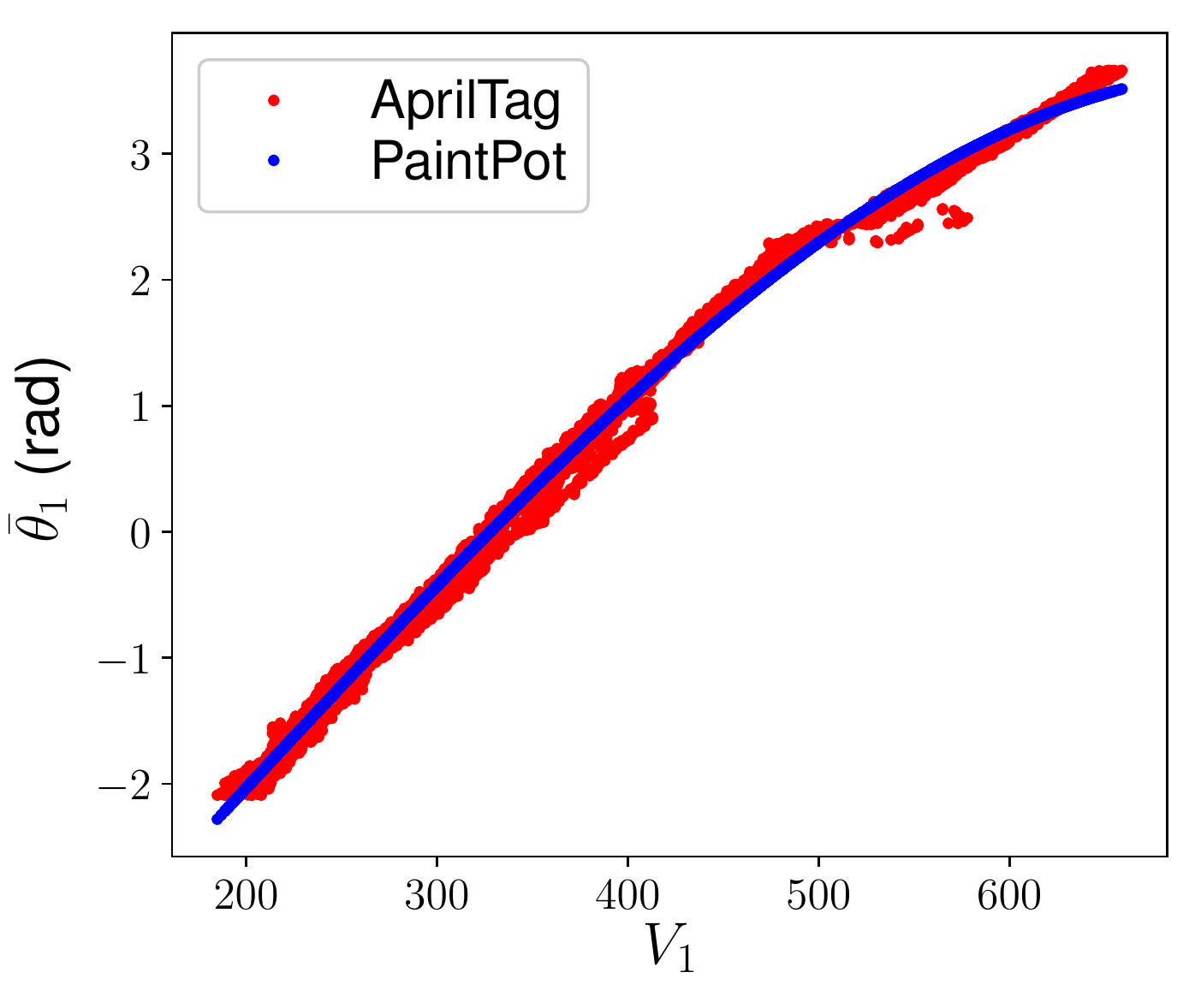}\label{fig:wiper1-experiment-measure}}
  \caption{Wheel PaintPot sensor characterization results: (a)
    $\bar{\theta}_0 = \bar{f}_0(V_0) = \num{-6.5012e-8}V_0^3 + \num{7.2912e-5}V_0^2
    - \num{1.1587e-2}V_0-3.4595$ and (b)
    $\bar{\theta}_1 = \bar{f}_1(V_1) = \num{-1.8511e-8}V_1^3 + \num{1.0419e-5}V_1^2
    + \num{1.4362e-2}V_1-5.1767$.}
\end{figure}

The data from both wipers for a wheel PaintPot is shown in
Fig.~\ref{fig:wiper0-experiment-data} and
Fig.~\ref{fig:wiper1-experiment-data} respectively. The segments
labeled by red color are useless, and the green segments are shifted
to the yellow segments to generate continuous functions to describe
the relationship between voltage and angular position. This sensor is
installed for PAN DOF on a SMORES-EP module. All our sensors are
painted manually, so the quality is not consistent with no guarantees on bounds. The sensor
characterization results for both wipers are shown in
Fig.~\ref{fig:wiper0-experiment-measure} and
Fig.~\ref{fig:wiper1-experiment-measure} respectively.

\begin{figure}[t]
  \centering
  \subfloat[]{\includegraphics[width=0.24\textwidth]{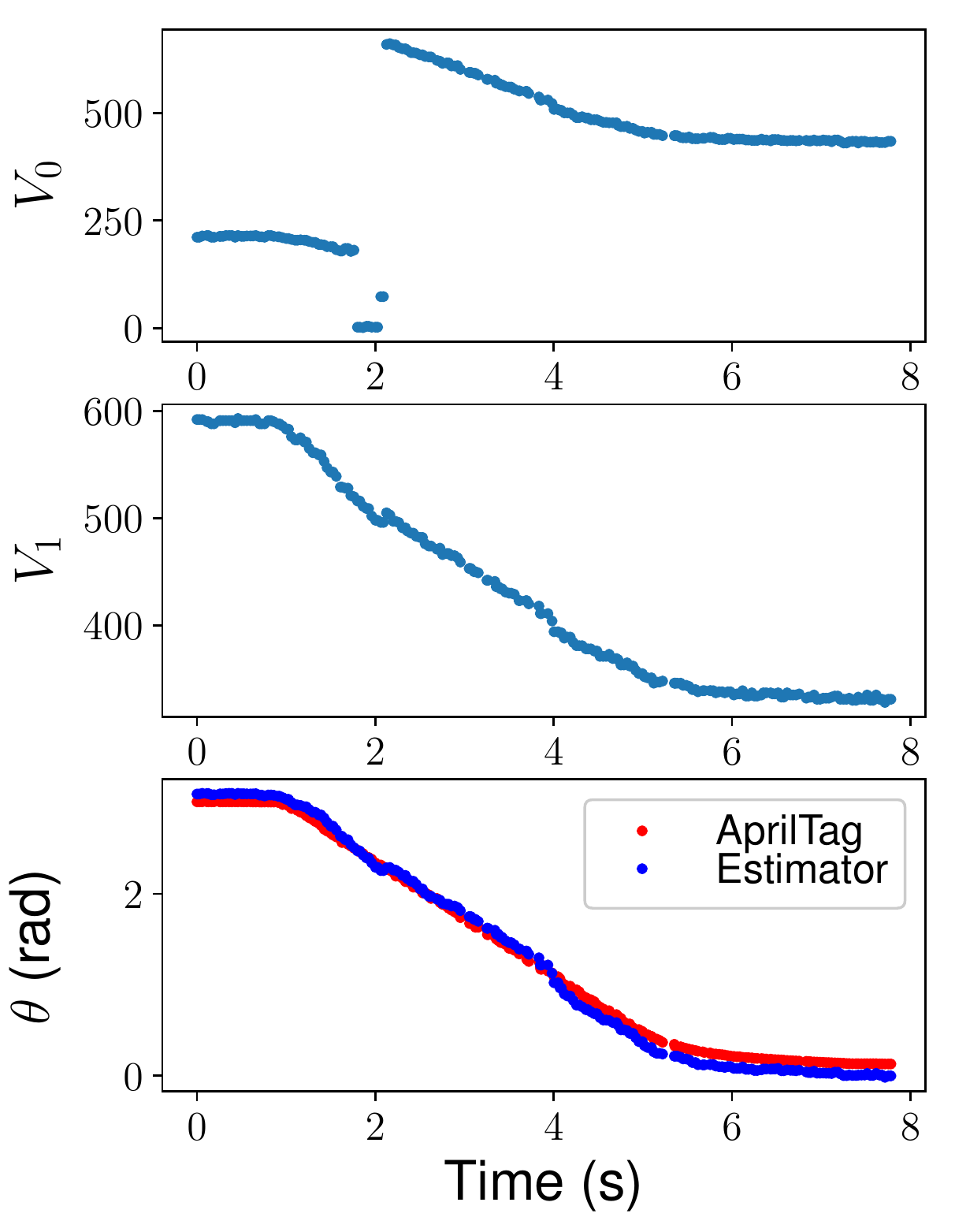}\label{fig:pan-experiment-0}}
  \hfil
  \subfloat[]{\includegraphics[width=0.24\textwidth]{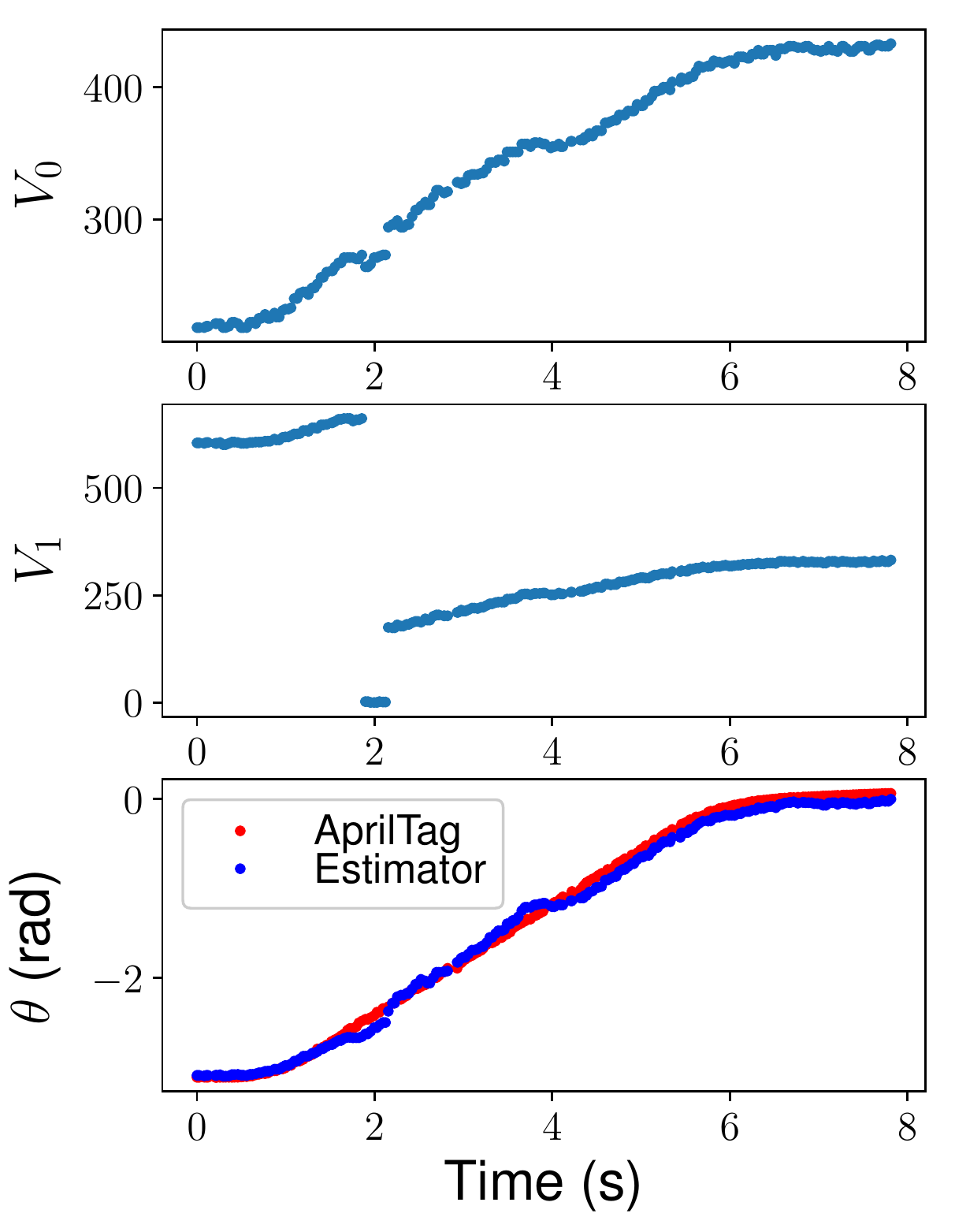}\label{fig:pan-experiment-1}}
  \caption{(a) Command PAN DOF to move from angular position
    \SI{\pi}{\radian} to \SI{0}{\radian}. (b) Command PAN DOF to move
    from angular position \SI{-\pi}{\radian} to \SI{0}{\radian}.}
\end{figure}

With our Kalman filter, we are still able to derive good estimation of
the angular position for the PAN DOF on this module. A simple
controller is used to command the PAN DOF to a desired position from
its current position along a fifth order polynomial trajectory.  The
first experiment commands the PAN DOF to angular position
\SI{0}{\radian} from \SI{\pi}{\radian} resulting in an average error of
\SI{0.0878}{\radian} as shown in Fig.~\ref{fig:pan-experiment-0}. In
this experiment, $V_0$ is not valid for a while because wiper 0
contacts the gap of the track. The second experiment commands the PAN
DOF to angular position \SI{0}{\radian} from \SI{-\pi}{\radian} with
average error being \SI{0.0698}{\radian} as shown in
Fig.~\ref{fig:pan-experiment-1}. In this experiment, $V_1$ is not
valid when wiper 1 contacts the gap of the track.

\begin{figure}[t]
  \centering
  \subfloat[]{\includegraphics[height=0.215\textwidth]{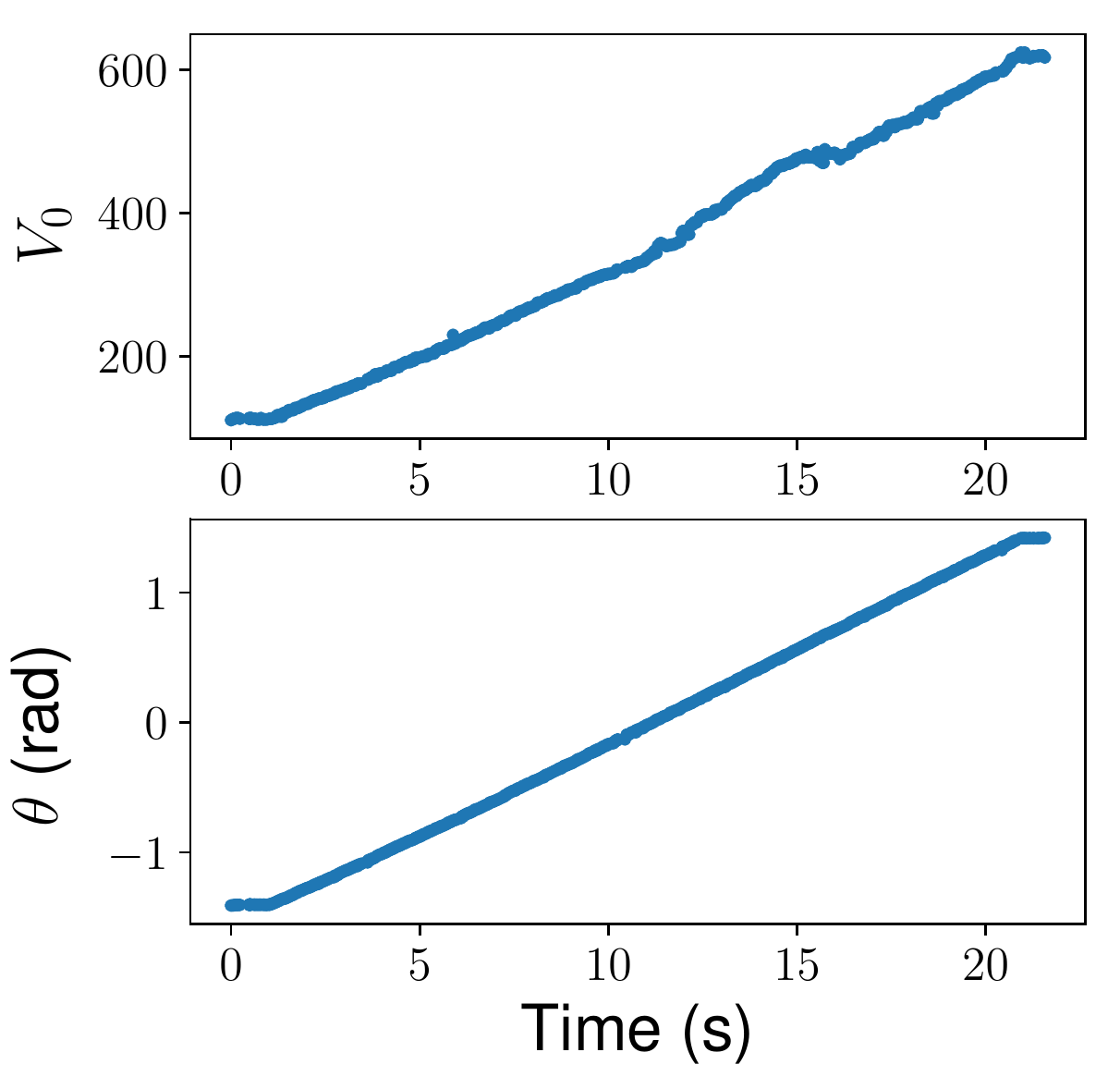}\label{fig:tilt-experiment-data}}
  \subfloat[]{\includegraphics[height=0.215\textwidth]{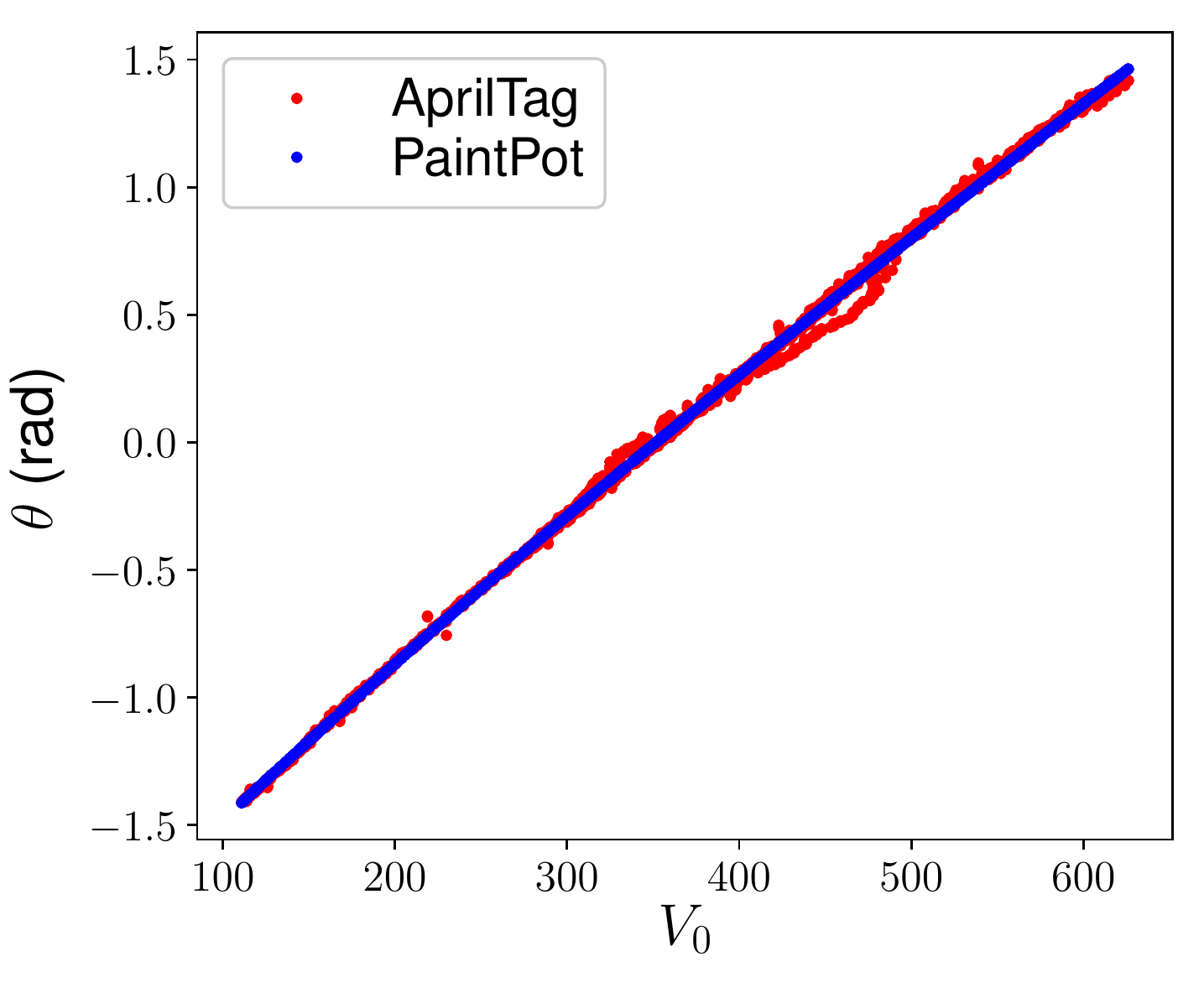}\label{fig:tilt-experiment-measure}}
  \caption{(a) Wiper data from \SIrange{-80}{80}{\degree} of
    a tilt PaintPot. (b) The characterization result $\theta = f(V_0) =
    \num{4.4674e-10}V_0^3-\num{1.5933e-6}V_0^2+\num{6.5369e-3}V_0-2.1117$}
\end{figure}

\begin{figure}[t]
  \centering
  \includegraphics[width=0.24\textwidth]{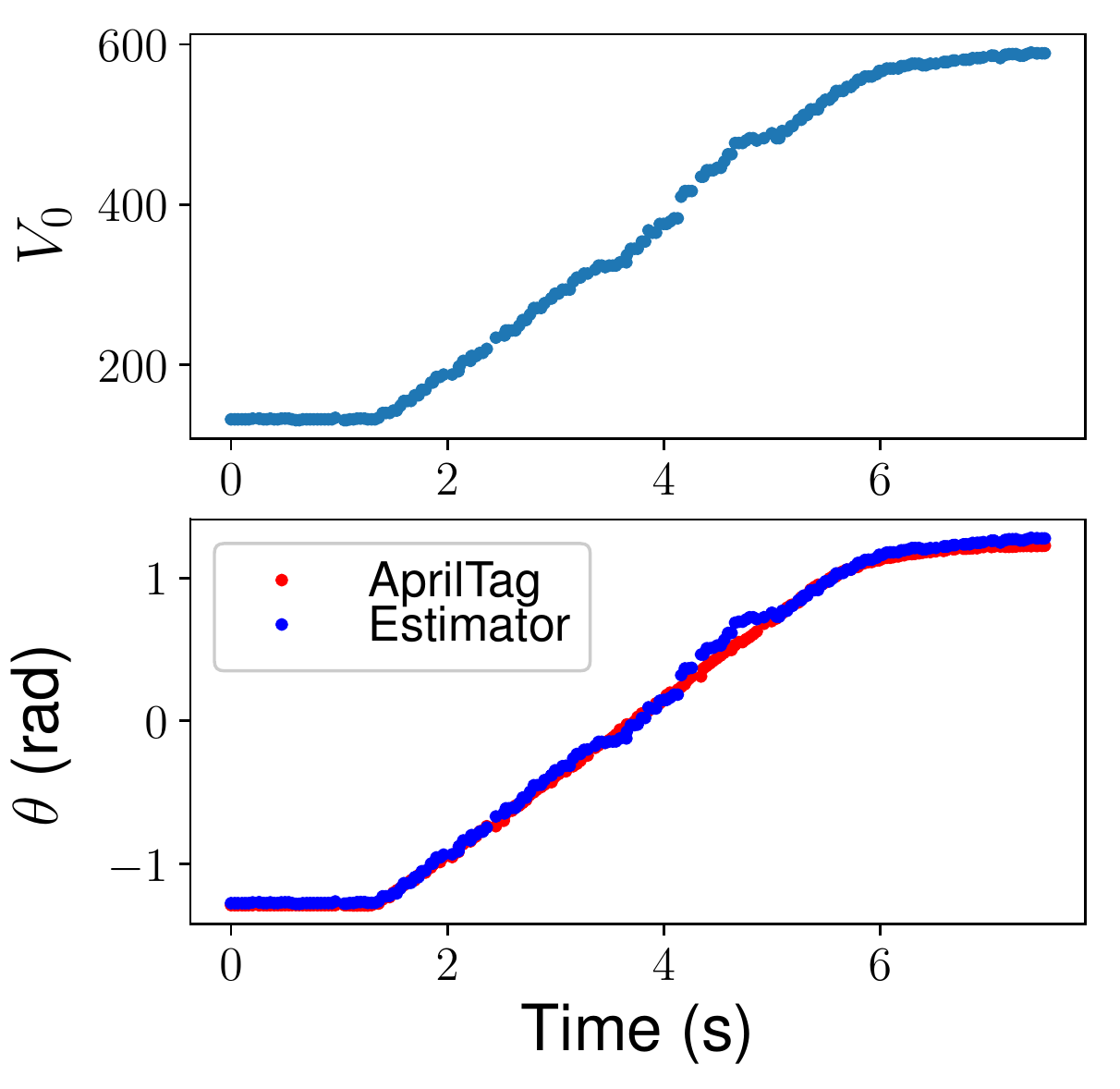}
  \caption{Command TILT DOF from angular position \SI{-1.3}{\radian} to \SI{1.3}{\radian}.}\label{fig:tilt-experiment}
\end{figure}

The data from the wiper for a tilt PaintPot is shown in
Fig.~\ref{fig:tilt-experiment-data} and the sensor characterization
result is shown in Fig.~\ref{fig:tilt-experiment-measure}. In the
experiment, this TILT DOF is commanded to traverse most of the
range. The result from our estimator is shown in
Fig.~\ref{fig:tilt-experiment} with an average error being around
\SI{0.0325}{\radian}.

\section{Conclusion}
\label{sec:conclusion}

In this paper, a complete low-cost and highly customizable position
estimation solution is presented, especially suitable for highly
space-constrained designs which is very common in modular robotic
systems. PaintPots are low-cost, highly customizable, and can be
manufactured easily by accessible materials and tools in low
quantities. For the SMORES-EP system, two different types of PaintPot
sensors are used, and a convenient automatic calibration approach is
developed using AprilTags. A modified Kalman filter is developed to
overcome the piece-wise nonlinearity of the sensors and some
experiments show the accuracy. The successful application of PaintPots
in SMORES-EP system shows that they can provide reliable position
information with simple hardware setup. PaintPots can be easily
adapted to and installed on a variety of systems, the not consistent
performance due to the manufacturing process can be resolved by our
characterization process which can be easily set up, and reliable
state estimation can be derived by our modified Kalman filter that can
be running on a low-cost microcontroller. The overall solution
provides a new position sensing technique for a wide range of
applications.

\begin{acknowledgment}
  We would like to thank Hyun Kim for experiments and hardware
  maintenance. This work was funded in part by NSF grant number
  CNS-1329620.
\end{acknowledgment}

%
% ...

\bibliographystyle{asmems4}

\bibliography{reference}



\end{document}